\documentclass[pdflatex,sn-mathphys-num]{sn-jnl}

\usepackage{graphicx}%
\usepackage{multirow}%
\usepackage{amsmath,amssymb,amsfonts}%
\usepackage{amsthm}%
\usepackage{mathrsfs}%
\usepackage[title]{appendix}%
\usepackage{xcolor}%
\usepackage{textcomp}%
\usepackage{manyfoot}%
\usepackage{booktabs}%
\usepackage{algorithm}%
\usepackage{algorithmicx}%
\usepackage{algpseudocode}%
\usepackage{tabularx} %
\usepackage{multirow} %
\usepackage{placeins} %

\usepackage{listings}%

\usepackage{pgfplots}
\pgfplotsset{compat=1.17} 
\usepackage{subcaption}

\theoremstyle{thmstyleone}%
\theoremstyle{thmstyletwo}%

\theoremstyle{thmstylethree}%

\begin{document}

\title[Article Title]{Solve it with EASE}


\author*[1]{\fnm{Adam} \sur{Viktorin}}\email{aviktorin@utb.cz}
\equalcont{All authors contributed equally to this research.}

\author*[1]{\fnm{Tomas} \sur{Kadavy}}\email{kadavy@utb.cz}

\author[1]{\fnm{Jozef} \sur{Kovac}}\email{j\_kovac@utb.cz}

\author[2]{\fnm{Michal} \sur{Pluhacek}}\email{mpluhacek@agh.edu.pl}

\author[1]{\fnm{Roman} \sur{Senkerik}}\email{senkerik@utb.cz}

\affil*[1]{\orgdiv{Faculty of Applied Informatics}, \orgname{Tomas Bata University in Zlin}, \orgaddress{\street{T. G. Masaryka 5555}, \city{Zlin}, \postcode{76001}, \country{Czech Republic}}}

\affil[2]{\orgdiv{Center of Excellence in Artificial Intelligence}, \orgname{AGH University of Krakow}, \orgaddress{\street{al. Mickiewicza 30}, \city{Krakow}, \postcode{30-059}, \country{Poland}}}


\abstract{This paper presents EASE (Effortless Algorithmic Solution Evolution), an open-source and fully modular framework for iterative algorithmic solution generation leveraging large language models (LLMs). EASE integrates generation, testing, analysis, and evaluation into a reproducible feedback loop, giving users full control over error handling, analysis, and quality assessment. Its architecture supports the orchestration of multiple LLMs in complementary roles—such as generator, analyst, and evaluator. By abstracting the complexity of prompt design and model management, EASE provides a transparent and extensible platform for researchers and practitioners to co-design algorithms and other generative solutions across diverse domains.}




\maketitle

\section{Introduction}
Computers have become deeply embedded in nearly every aspect of human activity over the past century. Almost as old is the ambition to automate the design of the very algorithms, codes, and programs that power these machines. For instance, the very origins of evolutionary computation are connected with the desire to have computers produce programs to solve problems \cite{Fogel1966}.

Traditionally, creating software is a complex and time-intensive task, demanding deep expertise. Thus, the central goal of automated algorithm design is to delegate the task of developing algorithms to the computer itself - reducing the need for expert knowledge and enabling users to generate effective solutions with minimal manual effort. In theory, efficient automated algorithm design should allow immense advances in numerous areas, and many promising results have been reported over the years \cite{kant2006understanding,francesca2014automode,stutzle2018automated,ma2025toward}

The latest significant step forward came with the resurgence of high-performance Large Language Models (LLMs), which have provided a powerful new platform for remarkable advances in automated algorithm design. 
To harness this power and steer it in the desired direction, researchers have once again turned to evolutionary algorithms to guide the algorithm creation process in an iterative way \cite{funsearch2024}.

This paper moves beyond prior efforts by introducing a fundamentally more versatile and modular framework for LLM-guided algorithm design that is not limited to any single application domain or guiding mechanism. The framework, titled EASE (Effortless Algorithmic Solution Evolution), is an open-source platform designed to enable researchers, practitioners, and the broader public to leverage the capabilities of LLMs for solving a wide range of tasks.

The development ofthe EASE framework is based on our long-term research into bio-inspired optimization algorithms, commonly known as metaheuristics or evolutionary/genetic algorithms \cite{bartz2014evolutionary}. Over the past decade, there has been criticism regarding the number of algorithms that claim to be new and metaphorically inspired \cite{sorensen_metaheuristicsmetaphor_2015, camacho2023exposing}. The academic community has called for more in-depth research into the behavior and execution of these algorithms, including their selection, (auto)configuration, and automated design \cite{camacho2023designing}. With the advancement of generative AI techniques, there has been significant progress in the area of automatic algorithm creation, including the configuration of algorithms according to user specifications.

Our first steps in this rapidly developing domain were taken in 2023 by investigating the capabilities of LLMs in selecting algorithmic components, reasoning, and designing a hybrid evolutionary algorithm \cite{pluhacek2023leveraging}. This initial study was followed by a focus on benchmarking suite selection and the performance-oriented design of specific adaptive functions for the Differential Evolution algorithm \cite{pluhacek2023investigating}. The promising results led to the idea of using an iterative approach instead of relying on manual interaction with the LLM.

The straightforward iterative feedback method, which instructed the LLM to improve performance, was explored in \cite{pluhacek2024critical}, along with a parametric setup for the LLM. The encouraging benchmarking results indicated that this simple approach could yield algorithms with better performance than one that placed third in the 2022 IEEE CEC competition \cite{cec2022}. This inspired further ideas and developments towards fully automated iterative feedback, using selected information to drive the LLM. The success of this approach was demonstrated in the automatic discovery of metaheuristics for optimizing real-world problems, such as time delay system control \cite{senkerik2025open}.  

At the same time, the first frameworks specifically designed for the automatic "evolution" of metaheuristics emerged, such as Large Language Model Evolutionary Algorithm (LLaMEA) \cite{van2024llamea}, Evolution of Heuristics (EoH) \cite{liu2024evolution}, and ReEvo \cite{ye2024reevo}. During our initial successes with metaheuristics development, we realized that this process could be generalized to a wide range of possible outputs. Therefore, we focused on developing a universal framework that is not exclusively aimed at metaheuristics.

Our goal was to create a universal and fully modular framework that allows the generation of algorithms, AI models, text, or even images. We also aimed to enhance the user experience, ensuring that interacting with this framework would be intuitive and straightforward. Users should find it easy to set up tasks and combine modules for testing, evaluation, and more.

The secondary objective was driven by the idea that this framework could bridge the gap between the field of AI and practitioners, making it easier for experts to work without needing in-depth knowledge of specific areas like optimization algorithms or machine learning models. By specifying the solved problem, input, desired output, evaluation conditions, defining the termination criteria, and providing the available data (if necessary), users should be able to achieve high-quality outputs.

These facts clearly define a targeted broader group of users, as the framework's user-friendliness means it can be used by researchers, algorithm and AI pipeline developers, domain experts, and practitioners alike.

\section{Background and Related Work}

LLMs have significantly advanced AI’s coding abilities, transforming them into capable programming assistants. Early developments like OpenAI's Codex (2021), the GPT-3-based model behind GitHub Copilot, demonstrated the translation of natural language into executable code \cite{yeadon2024comparison}. Successive benchmarks such as HumanEval and MBPP have shown rapid improvements, especially with iterative methods where LLMs plan, test, and refine their code. Notably, GPT-4 has substantially outperformed its predecessor GPT-3.5, achieving high accuracy and nuanced understanding.

Other proprietary LLMs also reflect this trend. Anthropic’s Claude series (Claude 2, 3.5, and 4) has steadily enhanced its programming and reasoning capabilities, at times outperforming GPT-4. Google’s Gemini 2.5 model (2025), featuring a vast context window of one million tokens, is optimized for handling extensive programming tasks in a single pass, thus excelling in complex software projects and algorithm design.

Open-source models, including Code Llama, DeepSeekCoder, and StarCoder2, now rival proprietary models on various coding benchmarks. These models benefit from community-driven adaptability, allowing fine-tuning for niche programming languages and specific use cases.

The substantial advances in coding proficiency since 2023—from basic script generation in GPT-3.5 to software development in current proprietary and open models—establish a foundation for iterative, automated algorithm discovery. Frameworks such as FunSearch \cite{funsearch2024} and LLaMEA \cite{van2024llamea} illustrate this potential, enabling LLMs to autonomously generate entire algorithmic structures, yielding innovative solutions to complex optimization problems. To mitigate challenges like hyperparameter tuning, hybrid approaches like LLaMEA-HPO \cite{van2024loop} integrate classical optimization techniques with LLM-driven algorithm design. Additionally, frameworks like BLADE \cite{van2025blade} and LLMatic \cite{nasir2024llmatic} have expanded automated algorithm generation to include benchmarking and neural architecture search.

Generative AI has also facilitated the development of AI research assistants such as Sakana AI \cite{lu2024ai} and Google's AI co-scientist \cite{gottweis2025towards}, automating the iterative scientific discovery process. In specialized application domains, frameworks like BrainBench \cite{luo2025large}, AlphaFold3 \cite{abramson2024accurate}, AlphaProteo \cite{zambaldi2024novo}, GenCast \cite{price2023gencast}, V-JEPA2 \cite{assran2025v}, and MedGemini \cite{saab2024capabilities} leverage generative AI for innovative research across neuroscience, biology, weather forecasting, robotics, and medicine.

These advancements collectively highlight the growing capability of LLMs to support iterative and automated discovery in software engineering and broader scientific research.

\section{Motivation}

The development of EASE was directly informed by the first experiments with automatically generated optimization algorithms  \cite{pluhacek2023leveraging, pluhacek2023DE_LLM, pluhacek2024critical}. Early prototypes using GPT‑3.5 and GPT‑4 demonstrated that LLMs could propose novel metaheuristics when given structured prompts and feedback. However, these experiments also revealed key limitations: the lack of a standardized evaluation loop, difficulties in maintaining reproducibility across iterations, and the absence of a modular structure to integrate different problem domains or testing environments. While frameworks like LLaMEA \cite{van2024llamea} and EoH \cite{liu2024evolution} addressed more or less parts of these challenges, they were often tied to specific benchmarking/evaluation tools or operated as closed, monolithic systems. Other approaches, such as FunSearch \cite{funsearch2024}, provided flexibility but required extensive computational budgets and lacked support for iterative refinement beyond single-function discovery. The closest alternative is Google's DeepMind AlphaEvolve \cite{novikov2025alphaevolve}, which may have similar features and functionalities; however, it operates as a closed system (not open-source) with limited limited overal reproducibility (user-selected SOTA algorithms, benchmarking results and capabilities).

These observations motivated the creation of the EASE framework. EASE was designed to generalize the lessons from early LLM-generated optimization research into a fully open, modular, and user-oriented platform. Its architecture supports a continuous \textit{test--evaluate--refine cycle}, ensuring that generated solutions can be iteratively improved and benchmarked in a reproducible way. Unlike domain-specific systems, EASE is intended to accommodate diverse applications: from metaheuristic algorithm design and automated code synthesis to integration with external evaluation pipelines. 

By embedding state-of-the-art LLM coding abilities within a flexible iterative workflow, EASE aims to provide researchers with a sustainable environment for exploring and accelerating algorithmic innovation. The framework bridges the gap between one-off LLM experiments and a systematic, extensible platform for automated discovery, enabling not only the automation of existing tasks but also the exploration of entirely new solution spaces. A major advantage of our framework is its full modularity, where, for example, with minimal effort (converting existing evaluations functions to the EASE interface), all of the engineering use cases (mentioned in the previous section) with specifically designed frameworks can also be solved with EASE. Moreover, any generator can also be used, as well as combinations of generators.

\section{EASE}

EASE is a modular and extensible framework designed to automate the iterative creation and refinement of algorithmic or generative tasks. Its core purpose is to harness the power and flexibility of large language models — or any other compatible generators — to co-design solutions based on a user-defined task specification. While LLMs are the default generators, EASE is not tied to them, thanks to its abstraction and interchangeable components.

At the heart of the framework lies the concept of a \textbf{Task}. A Task defines what is to be generated — whether it's an algorithm, an image, a piece of code, or any structured output — and specifies how the system should evaluate, iterate, and improve the solution.

The workflow begins with the user defining a \textbf{Task}, which serves as a modular specification of the problem to be solved. This includes prompts that guide the generator, optional behavioral constraints, and templates for iterative feedback. These prompts are fed into the \textbf{LLM connector} module, which manages communication with the selected generator and returns a proposed \textbf{solution}. The \textbf{Solution} module handles the expected output format, which can range from plain text to structured code, images, or other types.

The generated solution is then tested using the \textbf{Test}s module. The user defines these tests, which may contain syntax checks or logical validation. If an error is detected, the system loops back using an error-specific feedback prompt to guide the LLM toward correction. When the solution passes testing, it proceeds to the \textbf{Analysis} module, which can extract additional metadata or performance indicators and provide additional feedback into repetitive prompt. This is followed by the \textbf{Evaluation} module — one of the key components of EASE — which quantitatively or qualitatively assesses the quality of the solution. This evaluation can be problem-specific, such as benchmarking a generated algorithm or scoring image fidelity.

After evaluation, the system checks against a set of \textbf{Stopping conditions}. If any are satisfied (e.g., a fitness threshold is met, or a maximum number of iterations is reached), the \textbf{Task} ends. Otherwise, the loop continues, incorporating evaluator feedback to generate improved solutions.

Finally, upon termination, the \textbf{Statistics} module compares and summarizes the progress across all iterations, providing insights and visualizations of the solution evolution process.

\begin{figure}[h]
\centering
\includegraphics[width=1.0\textwidth]{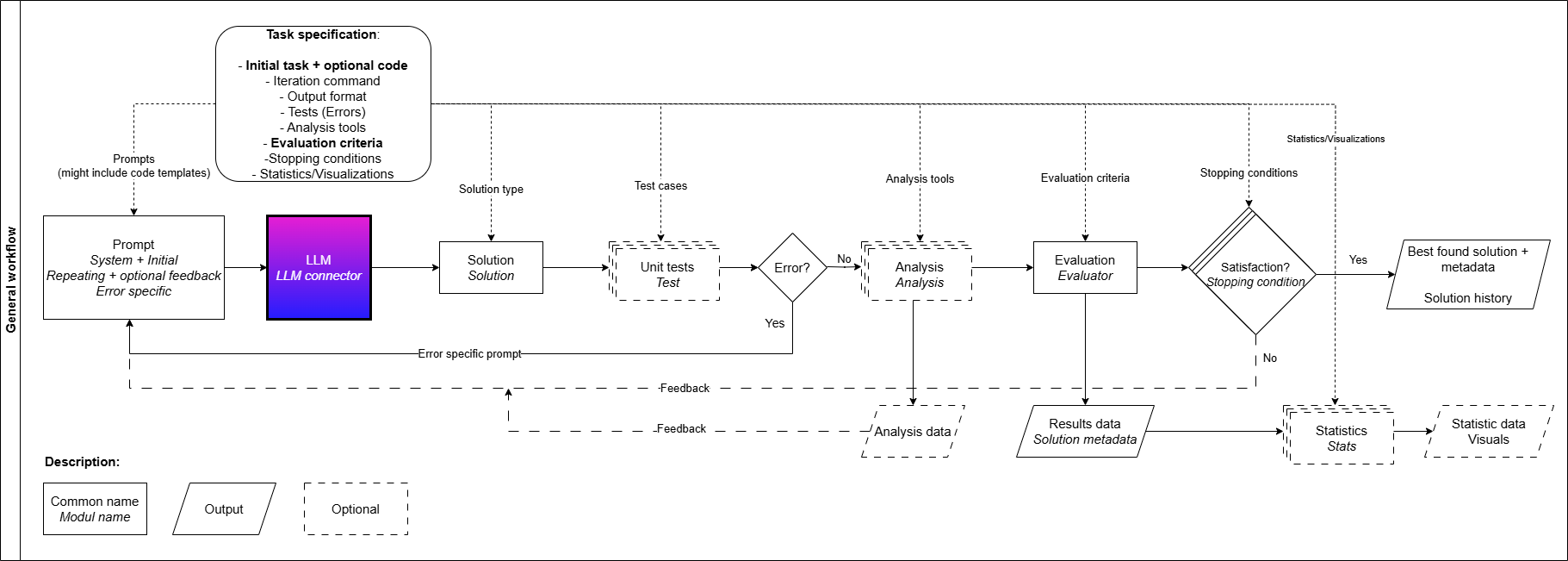}
\caption{EASE workflow.}\label{fig1}
\label{EASE:workflow}
\end{figure}

The entire process is orchestrated in a fully modular and replaceable way, enabling users to customize each step of the loop — from generation and testing to analysis and evaluation — depending on the nature of their problem domain. The full structure of the process is depicted in Fig.~\ref{EASE:workflow}.

The following subsections provide a detailed description of each individual module, including their functionality, modularity, and integration within the overall workflow.


\subsection{Prompts}

Before the iterative loop begins, each Task must define the initial communication with the generative agent — typically a large language model (LLM). This is done through a set of structured messages inspired by the OpenAI chat API format. Although not a modular component in the same sense as other parts of the framework (e.g., testing or evaluation modules), prompt design plays a foundational role in guiding the behavior and intent of the generator.

At the beginning of the task execution, two types of messages may be defined:

\begin{itemize}
    \item \textbf{Initial Prompt:} A user-defined message that describes the task to be solved. This serves as the entry point for the model and typically includes a clear specification of the desired output or behavior (e.g., “Generate an evolutionary algorithm for numerical optimization.”).
    
    \item \textbf{System Prompt:} An auxiliary message that sets the behavior or persona of the model. This concept is borrowed directly from the OpenAI chat paradigm, where the system prompt frames the model's identity or perspective (e.g., “You are a software engineer specialized in C programming.”).
\end{itemize}

These messages are sent once at the beginning of the Task, and the System message persists throughout the iterative process. These messages establish the context and tone of the conversation.

In addition to these, the framework supports an optional component called the \textbf{Repeating Prompt}. This is a user-defined collection of one or more messages that are automatically included at the beginning of each feedback prompt after a successful iteration. These repeating messages are intended to reinforce long-term instructions, preserve continuity, or emphasize goals across iterations.

The user can define not only the content of the repeating messages but also the \textbf{selection strategy} used to choose which one to insert at each step. The currently supported strategies include:
\begin{itemize}
    \item \textbf{Single:} Always use the same predefined message.
    \item \textbf{Random:} Select a message uniformly at random from a defined set.
    \item \textbf{Random Weighted:} Select a message at random using user-defined weights.
    \item \textbf{Circular:} Cycle through the list in order, looping back to the beginning when the end is reached.
\end{itemize}

This mechanism allows users to maintain persistent reinforcement or vary the guidance dynamically without manually rewriting the feedback messages during task execution.


\subsection{LLM Connector}

The LLM Connector is a central component of the EASE framework responsible for interfacing with the generative model. Its primary role is to deliver the structured messages (system prompt, initial prompt, and any feedback prompts) to the selected large language model and to return the generated response, which is then treated as the candidate solution.

While the default configuration assumes the use of an LLM (such as models provided by OpenAI, Anthropic, Meta, or HuggingFace), the connector is designed to be model-agnostic. Thanks to its modularity, any generator — including non-LLM models, traditional code generators, human-in-the-loop or rule-based systems — can be integrated, provided they adhere to the expected input-output communication protocol.

The connector handles:
\begin{itemize}
    \item \textbf{Conversation formatting:} Composing the prompt messages into a structured format suitable for the target model's API (e.g., OpenAI's chat-based schema).
    \item \textbf{Communication:} Sending the prompt to the model endpoint and receiving the output in a consistent, abstracted format.
    \item \textbf{Metadata capture:} Recording response details such as model identity, timing, used tokens and output metadata. In addition to textual responses, the connector can handle multimodal outputs, including images or videos, if such generation is supported and requested by the user.
\end{itemize}

The LLM connector supports optional conversational memory, configurable at the Task level. The user can specify the size of the message context — defined as the number of previous messages to be included at the start of each iteration. This setting can range from zero (stateless behavior) to an unbounded history, allowing for rich context accumulation when needed. This design enables EASE to adapt seamlessly across different generators and providers while remaining flexible for both minimal and memory-intensive tasks.

\subsection{Solution}

The Solution module captures and formalizes the output produced by the generative model during each iteration of the Task. While the LLM connector handles communication with the generator, the Solution module defines the expected structure, format, and type of the generated artifact.

The type of solution is user-defined and should correspond to the nature of the Task. Common output types include:
\begin{itemize}
    \item \textbf{Text:} Plain or structured natural language output.
    \item \textbf{Code:} Source code in a specified language (e.g., Python, C).
    \item \textbf{Image:} Visual content, either as base64-encoded data or external file references.
\end{itemize}

By explicitly specifying the expected output type, the Solution module enables downstream modules (such as testing and evaluation) to process the result appropriately. For example, if the Task expects a C function, the testing module may invoke a syntax checker or compile the code. If the output is an image, analysis tools may perform similarity or style evaluation.

In addition to raw content, the Solution module can store auxiliary metadata such as timestamps, iteration counters, or identifiers to track the evolution of solutions over time. The design of this module ensures that EASE remains agnostic to content type while maintaining structured access to generated artifacts for automated evaluation, comparison, and visualization.

\subsection{Test}
The Testing module is responsible for validating the generated Solution before it proceeds to the Analysis and Evaluation stages. It serves as an automated checkpoint to detect errors, inconsistencies, or violations of expected properties of the generated Solution early in the iteration loop.

Tests are fully user-defined and optional and reflect the specific requirements of the Task. Depending on the Solution type and problem domain, tests can range from basic syntax validation to domain-specific behavioral checks. Examples include:
\begin{itemize}
    \item \textbf{Syntax checking:} Ensuring that generated code compiles or parses correctly.
    \item \textbf{Unit tests:} Running predefined test cases on code to verify correctness.
    \item \textbf{Content checks:} Verifying that key elements or keywords are present in the output (e.g., specific function names, format constraints).
\end{itemize}

If a test fails, the system triggers an automatic feedback loop. A predefined feedback message corresponding to the detected error is selected and injected into the next prompt, allowing the generative model to correct or improve the Solution. This feedback mechanism is essential for enabling iterative refinement without manual intervention.

The Testing module is optional but strongly recommended for tasks where correctness, safety, or functional compliance is critical. It acts as a safeguard that filters unfit solutions before deeper analysis, helping to focus evaluation efforts only on viable candidates.

\subsection{Analysis}

The optional Analysis module is responsible for extracting additional information or metadata from the generated Solution before it proceeds to the evaluation phase. Unlike testing, which focuses on correctness or constraint satisfaction, analysis is aimed at capturing the contextual or structural information from the generated Solution.

The exact nature of the analysis is user-defined and highly dependent on the type of solution being generated. Examples include:

\begin{itemize}
    \item \textbf{Code:} Extracting function names, counting lines of code, measuring cyclomatic complexity, or checking usage of specific libraries.
    \item \textbf{Text:} Analyzing sentiment, counting tokens, or computing readability scores.
    \item \textbf{Images:} Measuring resolution, color distribution, or structural similarity to reference samples.
    \item \textbf{LLM generated feedback:} An analysis provided by a LLM accessed via a analysis-dedicated LLM connector.
    \item \textbf{Custom metrics:} Any user-specified feature extractor or post-processing step that provides auxiliary information.
\end{itemize}

The output of the analysis module is typically a dictionary of structured data associated with the solution. This data can be used by the Evaluator, feed back into the Repeating prompt, logged for debugging purposes, or visualized at the end of the Task in the Statistics module.

By separating analysis from evaluation, the framework enables users to compute informative metrics without tying them directly to fitness or quality scoring. This separation promotes modularity and supports a wide range of experimental and diagnostic workflows.

\subsection{Evaluator}

The Evaluator is one of the core components of the EASE framework. It is responsible for assessing the quality, fitness, or utility of a generated solution based on user-defined criteria. Together with the LLM Connector, it forms the central logic of every Task, guiding the iterative refinement process by providing quantitative or qualitative feedback after each iteration.

The evaluation process typically takes into account:
\begin{itemize}
    \item The generated \textbf{solution} itself,
    \item Any associated \textbf{analysis data},
    \item Optionally, \textbf{historical context} or \textbf{previous best solutions}.
\end{itemize}

Evaluators may implement a wide range of scoring strategies, depending on the domain and problem type:
\begin{itemize}
    \item \textbf{Numerical fitness:} Used in optimization settings (e.g., error on a benchmark function).
    \item \textbf{Heuristic scoring:} Rule-based evaluation of desirable properties (e.g., novelty, compactness, or compliance).
    \item \textbf{Comparative scoring:} Comparing the current solution to the previous best or to multiple alternatives.
    \item \textbf{Multi-objective evaluation:} Producing a vector of criteria to be later aggregated or used in Pareto-based selection.
\end{itemize}

In cases where quantitative evaluation is difficult to define — such as creative text generation or image synthesis — the Evaluator can also delegate the assessment to a language model. The LLM may be prompted to rate, rank, or critique the solution, effectively acting as a qualitative evaluator. This approach is particularly useful when subjective criteria, such as clarity, coherence, aesthetics, or novelty, are involved and cannot be easily captured through explicit heuristics or numeric scoring.

The Evaluator outputs a structured evaluation result — typically a numeric score, a dictionary of partial metrics, or a list of ranked solutions. This result is then used to:
\begin{itemize}
    \item Inform the feedback mechanism for the next iteration,
    \item Contribute to the final statistics upon Task termination.
\end{itemize}

Because it encapsulates the notion of “quality” within the Task, the Evaluator is often the most domain-specific module in the framework. It allows users to define what “good” means in the context of their problem — whether it is accuracy, creativity, performance, simplicity, or a combination of factors.

\subsection{Stopping condition}

The Stopping Conditions module determines when the Task should terminate. It acts as the control mechanism that decides whether the iterative generation and evaluation process should continue or halt based on user-defined criteria.

Stopping conditions are evaluated after each iteration and can be based on various signals derived from the solution, evaluation results, or overall Task progress. Several types of stopping conditions are supported:

\begin{itemize}
    \item \textbf{Maximum number of iterations:} Terminates the Task after a fixed number of total iterations.
    \item \textbf{Token usage limit:} Ends the Task once the cumulative number of tokens consumed by the generator reaches a specified threshold.
    \item \textbf{Maximum number of valid iterations:} Stops the Task after a defined number of successful (i.e., error-free) iterations, as validated by the Testing module.
    \item \textbf{Maximum number of consecutive invalid iterations:} Terminates the Task if a sequence of failed attempts exceeds a user-defined limit.
    \item \textbf{Score threshold:} Ends the Task once a solution reaches or surpasses a target evaluation score.
    \item \textbf{Time limit:} Stops execution if the overall runtime exceeds a predefined duration.
\end{itemize}

Multiple stopping conditions can be combined within a single Task. By default, the system terminates when \emph{any} of the defined conditions is satisfied, though more advanced combination strategies may be supported in future extensions.

The flexibility of this module allows EASE to adapt to both exploratory tasks — where open-ended iteration is preferred — and well-defined optimization problems, where convergence or success can be precisely detected.

\subsection{Statistics}

The Statistics module is responsible for aggregating and presenting information collected throughout the execution of the Task. Its main purpose is to provide users with insights into the iterative process, the quality and progression of generated solutions, and the overall performance of the Task.

During each iteration, EASE can store metadata including evaluation scores, analysis results, test outcomes, and solution characteristics. The Statistics module collects this data and processes it at the end of the Task to produce meaningful summaries.

Typical outputs of this module include:

\begin{itemize}
    \item \textbf{Iteration history:} A complete log of solutions, scores, errors, and feedback messages across all iterations.
    \item \textbf{Best solution summary:} The final solution selected according to the evaluation criteria, along with associated metadata and reasoning.
    \item \textbf{Score progression:} A visualization or list of how evaluation scores evolved over time.
    \item \textbf{Error distribution:} Statistics on test failures, common error types, or frequency of invalid outputs.
    \item \textbf{Token and runtime usage:} Reporting on resource consumption across the Task lifespan.
\end{itemize}

This module is particularly valuable for comparative analysis across different tasks or configurations, enabling reproducibility, tuning, and deeper understanding of the generative process. It also provides the foundation for post-task reporting and visualization in user-facing tools or dashboards.

\section{Case Studies / Examples}
\subsection{Text Generation: Iterative Story Rewriting with LLM Evaluation}

To showcase EASE's capability in refining natural language outputs, we present a use case focused on iterative creative writing. Specifically, the task was to generate increasingly original and engaging versions of the classic fairytale ``Little Red Riding Hood,'' with each iteration incorporating a novel twist. This example demonstrates how EASE can facilitate creative exploration through structured language model feedback loops, with one LLM acting as the generator and another serving as an autonomous critic.

\subsubsection{Task Setup}

The Task involved two LLMs: OpenAI \texttt{o1} was used as the Solution Generator, and Claude Sonnet 3.7 served as the Evaluator. The objective was to iteratively rewrite the fairytale using the following prompt configuration:

\begin{itemize}
    \item \textbf{System prompt:} \textit{You are a witty fairytale writer. You specialize in short stories with unexpected twists.}
    \item \textbf{Initial prompt:} \textit{Write a short version of Little Red Riding Hood with an unexpected twist.}
    \item \textbf{Repeating prompt:} \textit{Introduce a new and better twist to the original story of Little Red Riding Hood.}
    \item \textbf{Feedback prompt (Evaluator):} \textit{Evaluate the quality of the original twist in the Little Red Riding Hood story and provide feedback in the following template: \\
    \texttt{Last twist evaluation:} \\
    \texttt{Rating: \{value\} note - rating from 1 to 10, 1 being the worst, 10 being the best} \\
    \texttt{Suggestion: \{text\} note - suggestion for a better story twist}}
\end{itemize}

This Task was configured to run for four iterations. At each step, the newly generated twist was evaluated numerically and qualitatively by the Evaluator, whose feedback was injected into the next generation loop.

\subsubsection{Iteration and Feedback Cycle}

Each iteration of the Task proceeded as follows:

\begin{enumerate}
    \item A new short story twist was generated using the current prompt context.
    \item The resulting story was passed to the Evaluator LLM.
    \item The Evaluator produced a structured critique consisting of a numerical score and a concrete suggestion for improvement.
    \item The suggestion was embedded as feedback into the next generation prompt.
\end{enumerate}

The iterative rewriting of Little Red Riding Hood explored various subversions of the traditional narrative. The first version introduced a gentle, poetry-reciting Wolf who co-ran a bakery with Grandma. The second added emotional depth by revealing the Wolf as a childhood companion secretly cherished by Grandma. The third emphasized guardianship, portraying the Wolf as Red’s lifelong protector. The final version reversed expectations by casting the Huntsman as the true threat, with the Wolf saving Red from harm. In each step, the Evaluator LLM assessed narrative coherence and originality, offering targeted suggestions that steered the next iteration toward greater emotional complexity and thematic nuance.

\subsubsection{Results and Discussion}

This case study demonstrates the efficacy of using different LLM roles within EASE—one for solution generation, another for autonomous evaluation. The structure allowed for creative evolution without requiring any human-in-the-loop oversight. The generated stories became progressively more inventive while maintaining narrative coherence, as shaped by the Evaluator's ongoing feedback.

However, the case also illustrates certain limitations. The quality ratings, though structured, remained subjective and varied slightly depending on how interpretive the Evaluator was in parsing the generator’s intent. Despite these nuances, the overall loop achieved continuous refinement across iterations, producing engaging variants with minimal setup.

This use case highlights EASE’s potential for facilitating autonomous creative workflows, especially in settings where subjective judgment must be encoded as structured feedback. It further validates the modularity of the Evaluator role—even in abstract or literary domains where quantitative scoring is inherently challenging.

The complete story evolution with feedback can be found in Appendix~\ref{secA1} and an example of one story/feedback pair is shown below:

\begin{quote}
\itshape
\textbf{Story:} \\
Little Red Riding Hood walked the forest path alone, her crimson cape bright as a flame in the shady green. Every so often, she sensed a presence behind the trees—soft rustling, gentle eyes glinting in the dark. She hurried on, never quite seeing who (or what) followed her.

At Grandma’s cottage, she felt immense relief… until Grandpa’s old rifle, which hung above the mantel, went off with a startling bang. Startled, Red rushed outside and spotted a wolf sprinting away, something clutched in his mouth. Heart hammering, she chased him into a clearing. Ready for the worst, she confronted him.

Except instead of snarling, the Wolf cautiously approached, dropping a delicate bundle of wildflowers at her feet. “These were Grandma’s favorites,” he said softly. His voice was hoarse and trembling with emotion.

Red blinked in astonishment. “You can talk?”

He nodded. “Your Grandma once saved my life when I was just a pup. I’ve been watching over you ever since you started walking these woods. Didn’t you ever notice that you’ve never been harmed on your journeys here?”

Suddenly, Red remembered the many times she had felt safe when stepping on shadowy paths and half-suspected someone was guiding her. “That was… you?” she asked hesitantly.

“I guarded you from real dangers,” the Wolf whispered. “Even when you were too scared to see, I was there.”

Before Red could respond, Grandma appeared, smiling. “I was testing both of you,” she confessed, hugging them both at once. “You, to see if your bravery could match your kindness… and Wolf, to ensure that your protective spirit is as loyal as ever.”

Red realized then: the forest’s spine-tingling rustles, those golden eyes watching from a distance, had never meant her any harm. They were a silent promise to keep her safe. And so, the cloak she wore wasn’t just a beacon for danger—it was a symbol of trust and unsaid gratitude between a granddaughter, a grandmother, and an unexpected guardian.
\end{quote}

\begin{quote}
\itshape
\textbf{Last twist evaluation:} \\
Rating: 7 Suggestion: The twist showing the wolf as a protective guardian rather than a threat is creative and heartwarming, effectively subverting reader expectations. However, it could be strengthened by addressing a few elements: First, the rifle going off feels disconnected from the plot - why did it fire, and how does this relate to Grandma's "test"? Second, Grandma's revelation that it was all a test feels somewhat contrived. Consider instead having the wolf save Red from a genuine threat (perhaps another predator or human danger), proving his guardianship in action rather than explanation. This would create a more organic revelation of the wolf's true nature and provide a more compelling emotional payoff while maintaining the core subversion of the traditional villain becoming a protector.
\end{quote}

\subsection{Algorithm Generation: 2048 Game Solver via Iterative LLM Refinement}

To demonstrate the capabilities of EASE in synthesizing algorithmic solutions, we present a concrete case study involving the development of a solver for the 2048 game. This game, while conceptually simple, poses a rich combinatorial challenge. It requires sequential decision-making under uncertainty, balancing short-term gains (e.g., combining tiles for immediate score boosts) with long-term strategic considerations (e.g., tile placement optimization to avoid grid locking). The task is especially well-suited for demonstrating the iterative generation, testing, and refinement capabilities of the EASE framework.

\subsubsection{Problem Specification and Prompt Design}

In the initial phase, a Task was created within EASE with clearly defined structural and functional constraints. The core requirement was to implement a Python function:

\begin{verbatim}
def move(grid: np.array, score: int) -> str:
    # Returns one of: 'left', 'right', 'up', or 'down'
    [Algorithm body]
    return direction
\end{verbatim}

This interface encapsulates a typical decision point in the 2048 game, where the agent receives the current $4\times4$ game grid (with zeros representing empty tiles) and the current game score. The goal is to select the most promising direction for the next move within a strict execution time constraint of 5 seconds.

The \textit{system prompt} contextualized the generative model as an expert Python developer and AI researcher. It instructed the model to leverage strategies such as heuristic scoring, lookahead simulation, and knowledge of tile-merging dynamics to guide its algorithmic decisions. The \textit{initial prompt} reiterated the task objective while imposing strong constraints on interface adherence, execution time, and coding completeness.

\subsubsection{Test and Evaluation Criteria}

The task incorporated a multi-stage validation pipeline:

\begin{itemize}
    \item \textbf{Syntax Verification:} The generated code was checked for correct Python syntax using \texttt{ast} parsing and static analysis tools.
    \item \textbf{Import Validation:} Only standard libraries and \texttt{numpy} were permitted; invalid imports triggered a rejection.
    \item \textbf{Functional Consistency:} A custom 2048 simulation engine tested whether the proposed move was valid under the current grid state. Invalid moves or decisions that resulted in no grid change were penalized.
    \item \textbf{Timing Constraint:} The move function was required to execute within the 5-second runtime constraint.
\end{itemize}

Evaluation was performed using three primary metrics:
\begin{enumerate}
    \item \textbf{Cumulative score} achieved during simulated gameplay.
    \item \textbf{Maximum tile value} reached during each game.
    \item \textbf{Number of valid moves} executed before termination.
\end{enumerate}

Each solution was tested over a series of simulated games using randomized seeds. Aggregate statistics were logged for every iteration, including score distributions, solver runtime, and move legality ratios.

\subsubsection{Iteration and Feedback Cycle}

EASE conducted five full iterations, guided by the maximum valid iteration stopping condition. Each iteration followed the same structured loop:

\begin{enumerate}
    \item A new version of the solver was generated using the configured prompts.
    \item The solution was tested across the defined criteria.
    \item If the code failed any test (e.g., due to syntax errors or move logic issues), it was rejected and annotated with corrective feedback.
    \item If the code passed all tests, it was evaluated using the metrics above, and performance statistics were recorded.
    \item Feedback — derived from performance degradation, lack of tile merging, or overuse of a single direction — was synthesized into a contextual prompt and submitted to the generator for the next iteration.
\end{enumerate}

In the first iteration, the generated solver relied on a simple heuristic evaluation combining corner-weighted tile positions, empty cell count, and a basic monotonicity approximation. This approach encouraged placing high-value tiles in the top-left and maintaining open space, but it lacked any foresight. As a result, the solver often made locally optimal but strategically poor decisions, leading to premature stagnation near grid edges.

The second iteration introduced a basic expectimax strategy with a fixed search depth. This was the first time the solver began simulating both player actions and random tile spawns, allowing it to select moves with higher expected long-term utility. Although the evaluation function remained relatively simple, the integration of forward simulation significantly improved its decision-making robustness.

In the third iteration, iterative deepening was added to the expectimax framework. The solver adapted its search depth dynamically based on runtime constraints, while the evaluation function was expanded to include smoothness (penalizing abrupt tile changes) and refined monotonicity scoring. This produced more consistent performance across varied game states and ultimately yielded the highest-performing solver among all iterations.

The fourth iteration focused on refining the evaluation strategy by introducing an advanced corner-weighted "snake pattern" heuristic and explicit merge detection. Additionally, basic move ordering was introduced to improve early pruning during search. While these changes improved tile arrangement aesthetics and local stability, they did not translate into better overall scores or tile progression than the third iteration.

The fifth and final iteration introduced a Monte Carlo-style fallback using randomized rollouts to break ties between similarly evaluated moves. While this made the solver slightly more adaptive in ambiguous states, the added stochastic component did not outperform the more consistent, deterministic approach of the third iteration.

\subsubsection{Results and Insights}

The final solver (iteration 3) achieved high performance, winning all 5 games and yielded the following statistics:

\begin{itemize}
    \item \textbf{Average score: 45,715.2} 
    \item \textbf{Average maximum tile:} 2,867.2
    \item \textbf{Game completion rate:} Reaching at least a 2048 tile in all 5 simulations
\end{itemize}

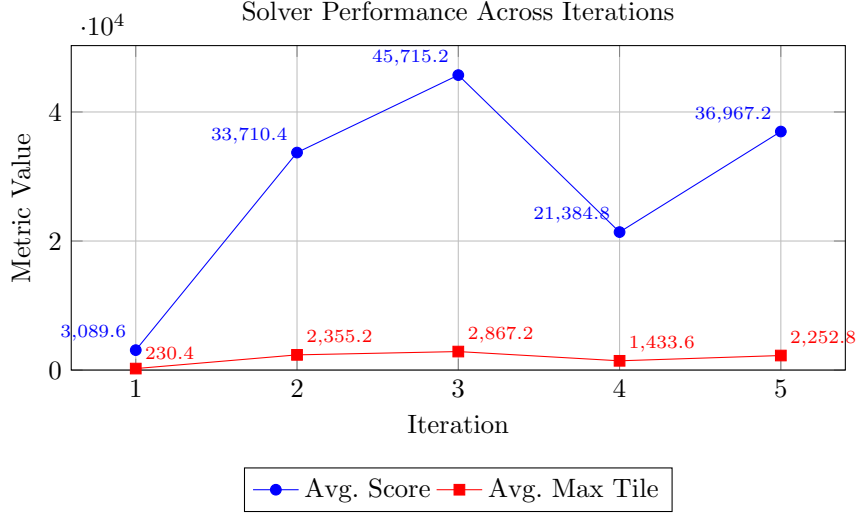
\begin{figure}[!ht]
    \centering
    \begin{tikzpicture}
        \begin{axis}[
            width=0.9\textwidth,
            height=0.45\textwidth,
            xlabel={Iteration},
            ylabel={Metric Value},
            ymin=0,
            xtick={1,2,3,4,5},
            legend style={at={(0.5,-0.3)}, anchor=north, legend columns=2},
            grid=both,
            title={Solver Performance Across Iterations}
        ]
        \addplot[
            color=blue,
            mark=*,
            nodes near coords,
            every node near coord/.append style={
                font=\footnotesize,
                anchor=south east,
                color=blue
            }
        ] coordinates {
            (1, 3089.6)
            (2, 33710.4)
            (3, 45715.2)
            (4, 21384.8)
            (5, 36967.2)
        };
        \addlegendentry{Avg. Score}

        \addplot[
            color=red,
            mark=square*,
            nodes near coords,
            every node near coord/.append style={
                font=\footnotesize,
                anchor=south west,
                color=red
            }
        ] coordinates {
            (1, 230.4)
            (2, 2355.2)
            (3, 2867.2)
            (4, 1433.6)
            (5, 2252.8)
        };
        \addlegendentry{Avg. Max Tile}
        \end{axis}
    \end{tikzpicture}
    \caption{Progression of solver performance over five valid EASE iterations. Each point is labeled with the corresponding value for both average game score and average maximum tile.}
    \label{fig:2048-curve}
\end{figure}

\begin{figure}[!ht]
    \centering
    \includegraphics[width=0.9\textwidth]{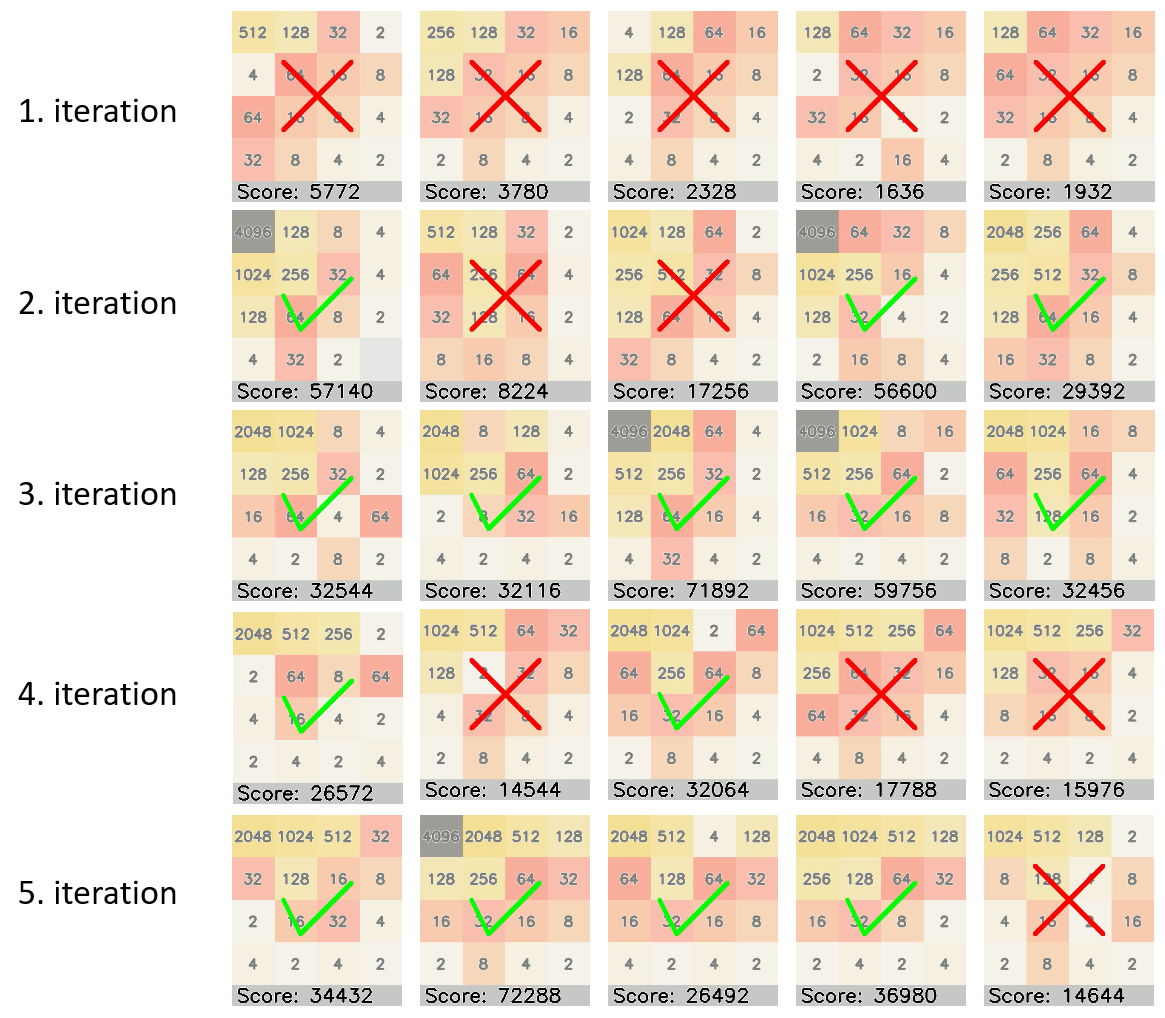}
    \caption{Example visualizations of final grid states of every run in each iteration.}
    \label{fig:2048-grid}
\end{figure}

Figure~\ref{fig:2048-curve} shows the progression of score and tile metrics across iterations, while Figure~\ref{fig:2048-grid} provides example visualizations of final grid states of every run in each iteration.

\subsubsection{Discussion}

This case study highlights several key strengths of the EASE framework:

\begin{itemize}
    \item \textbf{Prompt Engineering Integration:} The system and initial prompts effectively constrained the solution space, enabling the LLM to generate functional and compliant code early in the process.
    \item \textbf{Autonomous Correction:} The built-in testing module allowed for immediate rejection and correction of invalid code, making human intervention unnecessary during the refinement cycle.
    \item \textbf{Quantitative Feedback Loops:} By incorporating custom evaluators, EASE transformed a loosely defined goal ("solve 2048") into a quantitative, iterative optimization task.
    \item \textbf{Multi-objective Adaptation:} The solver evolved to balance tile merging efficiency, game longevity, and score maximization without explicit hand-tuning.
\end{itemize}

Overall, this example serves as a proof-of-concept that EASE can successfully navigate a constrained algorithm design space and yield performant solutions through structured LLM iteration.

The generated source code of each solver is included in Appendix~\ref{secA2}. 

\subsection{Image Generation: Guided Expansion of a Scientific Group Portrait}

To further demonstrate the flexibility of EASE across domains, we applied it to a multimodal design task: generating a scientifically themed group portrait via iterative visual augmentation. This example illustrates how language-based evaluators can guide a generative image model to build complex visual compositions in a step-by-step manner, mimicking curated content development.

\subsubsection{Task Setup}

The generative component of this Task used \texttt{DALL·E 3} as the image generator. The evaluation loop was driven by \texttt{OpenAI Vision o1}, tasked with analyzing each generated image and suggesting which historical scientist to add next. The system was configured to execute six iterations, where each iteration was designed to expand the existing group by adding one new figure.

The Task prompt configuration was as follows:
\begin{itemize}
    \item \textbf{System prompt:} \textit{You are an AI agent specialized in drawing realistic photo-like depictions of existing people.}
    \item \textbf{Initial prompt:} \textit{Draw portrait photo of a small group of famous scientists.}
    \item \textbf{Repeating prompt:} \textit{Keep the existing people and add one more person to the image.}
    \item \textbf{Feedback prompt (Evaluator):} \textit{Analyze the picture of a group of famous scientists and specify which person should be added to the group. Choose one of the existing famous scientists. Reply with a specification of the person and its proposed placement in the picture.}
\end{itemize}

\subsubsection{Test and Evaluation Criteria}

Unlike the algorithmic case, the evaluation here relied on qualitative analysis and semantic feedback. The Evaluator’s task was to:
\begin{itemize}
    \item Detect how many people were depicted and whether previous figures persisted.
    \item Recommend a well-known scientist who had not yet appeared.
    \item Suggest an appropriate placement that complemented the composition.
\end{itemize}

The effectiveness of each iteration was assessed by the clarity and relevance of additions, preservation of previous visual elements, and increasing thematic depth.

\subsubsection{Iteration and Feedback Cycle}

Each iteration followed the same structured loop:
\begin{enumerate}
    \item The current image was analyzed by the Evaluator module.
    \item Natural language feedback was generated, suggesting a new scientist and their position in the composition.
    \item This feedback was appended to the base repeating prompt and passed to the image generator.
    \item The newly generated image was evaluated for visual coherence and integration of the suggested change.
\end{enumerate}

Below is an example of feedback from the Evaluator after one of the intermediate iterations:

\begin{quote}
\itshape
Keep the existing people and add one more person to the image. One way to round out this group is to include Ada Lovelace, widely regarded as a pioneering figure in the development of computer science. You could position her just behind and to the left of the central figure in the back row, so that she fits naturally into the existing formation. Clad in a period-appropriate gown but wearing a lab coat to match the others, Lovelace would bring a computational and mathematical dimension to the group, reflecting the breadth of scientific fields represented.
\end{quote}

\subsubsection{Results and Insights}

The images evolved from a simple, studio-style four-person portrait to increasingly crowded and dynamic scenes, with up to a dozen individuals in diverse scientific settings. Early iterations were relatively controlled, with recognizable group composition and attire continuity. However, due to token limitations in the prompt, only the most recent user-assistant exchange and system instruction were preserved in each iteration. This led to notable drift: characters would change appearances, disappear, or be reintroduced inconsistently.

\begin{figure}[!ht]
    \centering
    \begin{subfigure}[t]{0.3\textwidth}
        \includegraphics[width=\textwidth]{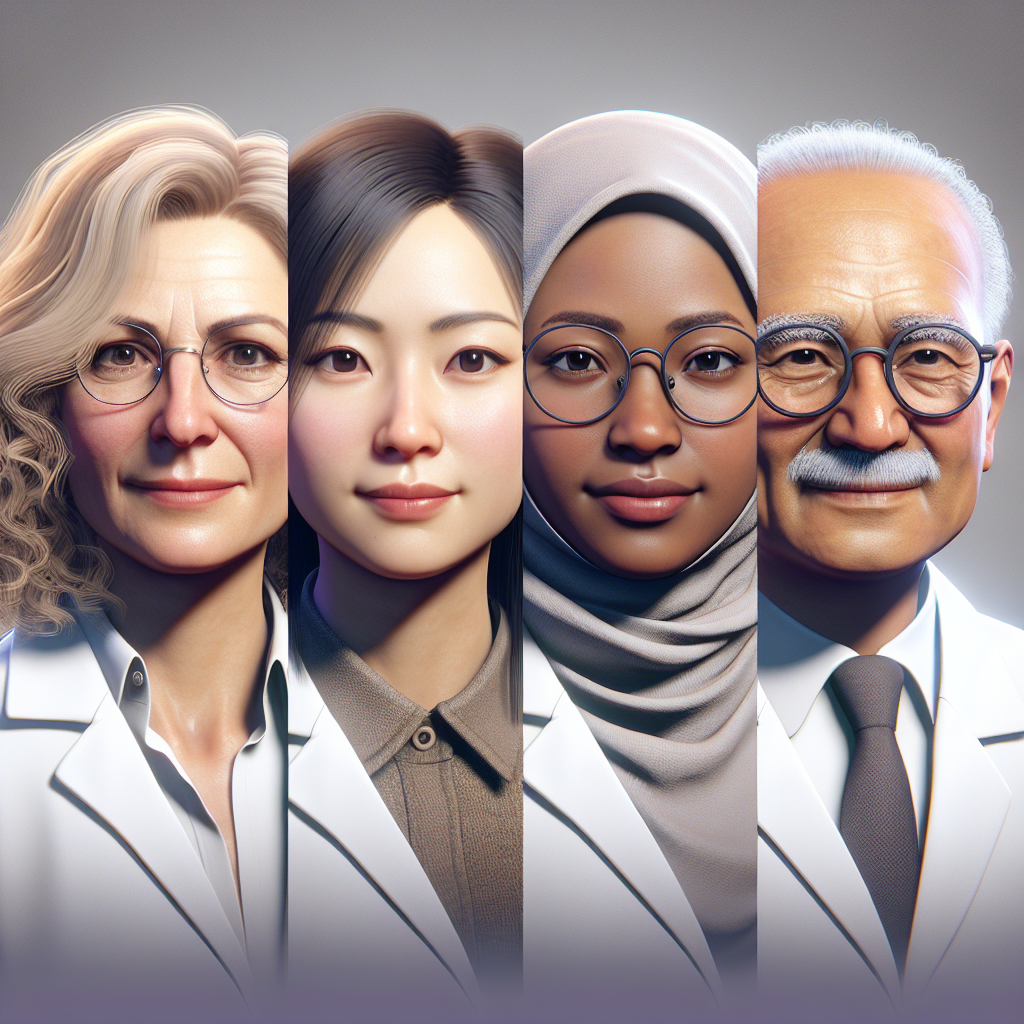}
        \caption*{Iteration 1}
    \end{subfigure}
    \hfill
    \begin{subfigure}[t]{0.3\textwidth}
        \includegraphics[width=\textwidth]{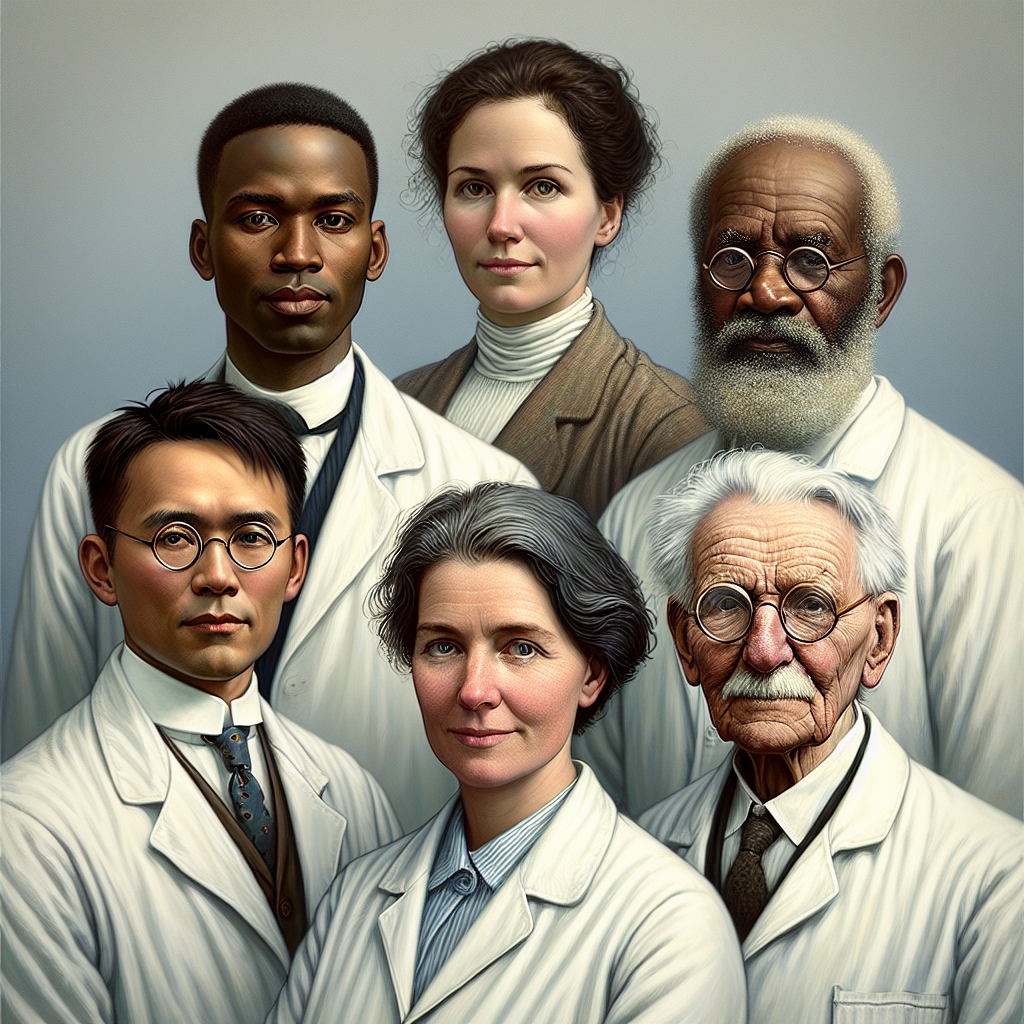}
        \caption*{Iteration 2}
    \end{subfigure}
    \hfill
    \begin{subfigure}[t]{0.3\textwidth}
        \includegraphics[width=\textwidth]{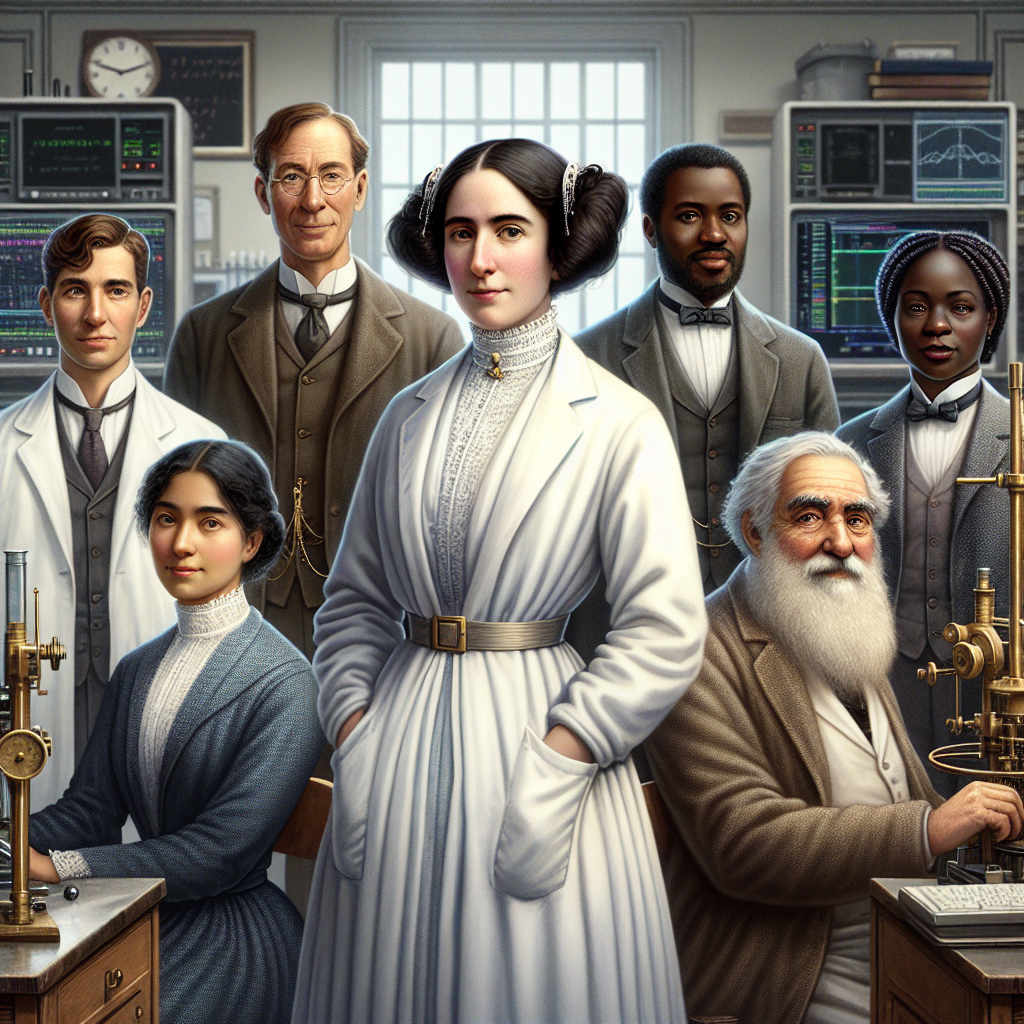}
        \caption*{Iteration 3}
    \end{subfigure}

    \vspace{1em}

    \begin{subfigure}[t]{0.3\textwidth}
        \includegraphics[width=\textwidth]{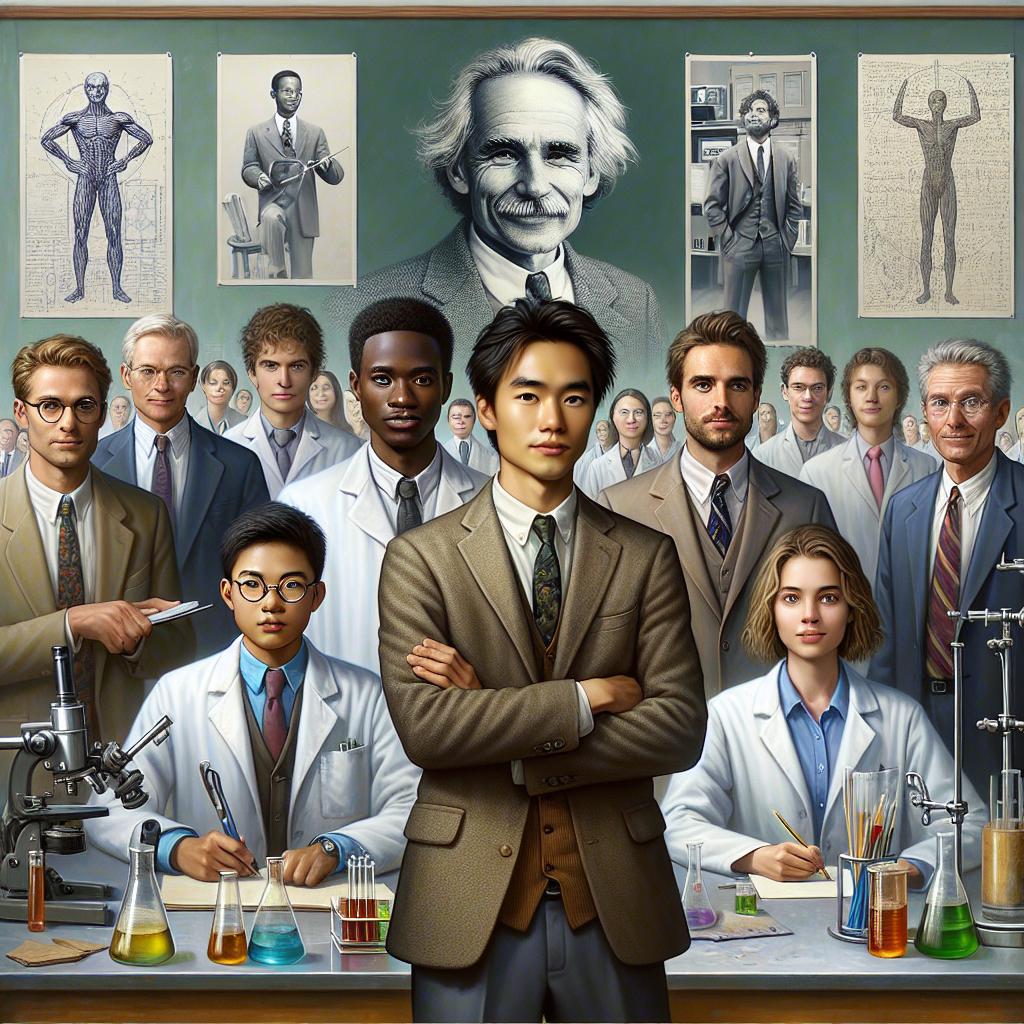}
        \caption*{Iteration 4}
    \end{subfigure}
    \hfill
    \begin{subfigure}[t]{0.3\textwidth}
        \includegraphics[width=\textwidth]{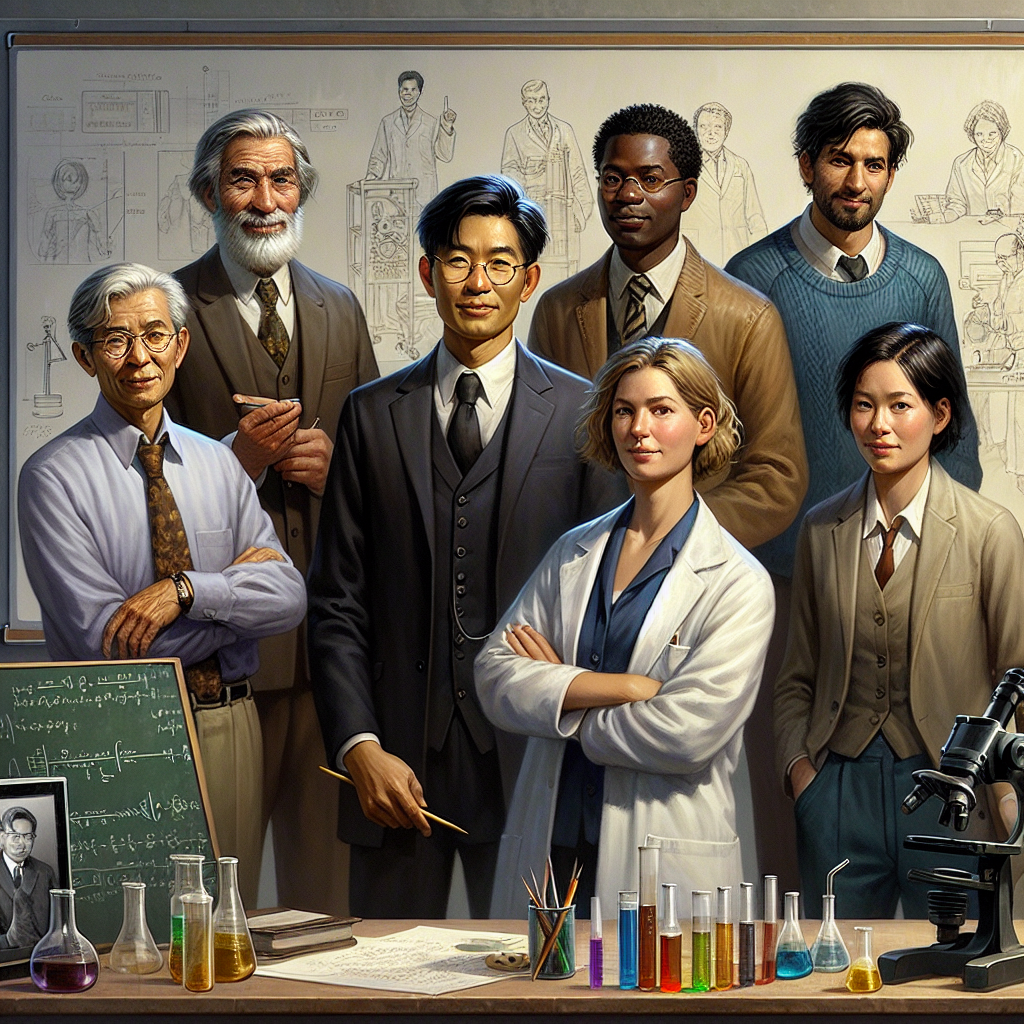}
        \caption*{Iteration 5}
    \end{subfigure}
    \hfill
    \begin{subfigure}[t]{0.3\textwidth}
        \includegraphics[width=\textwidth]{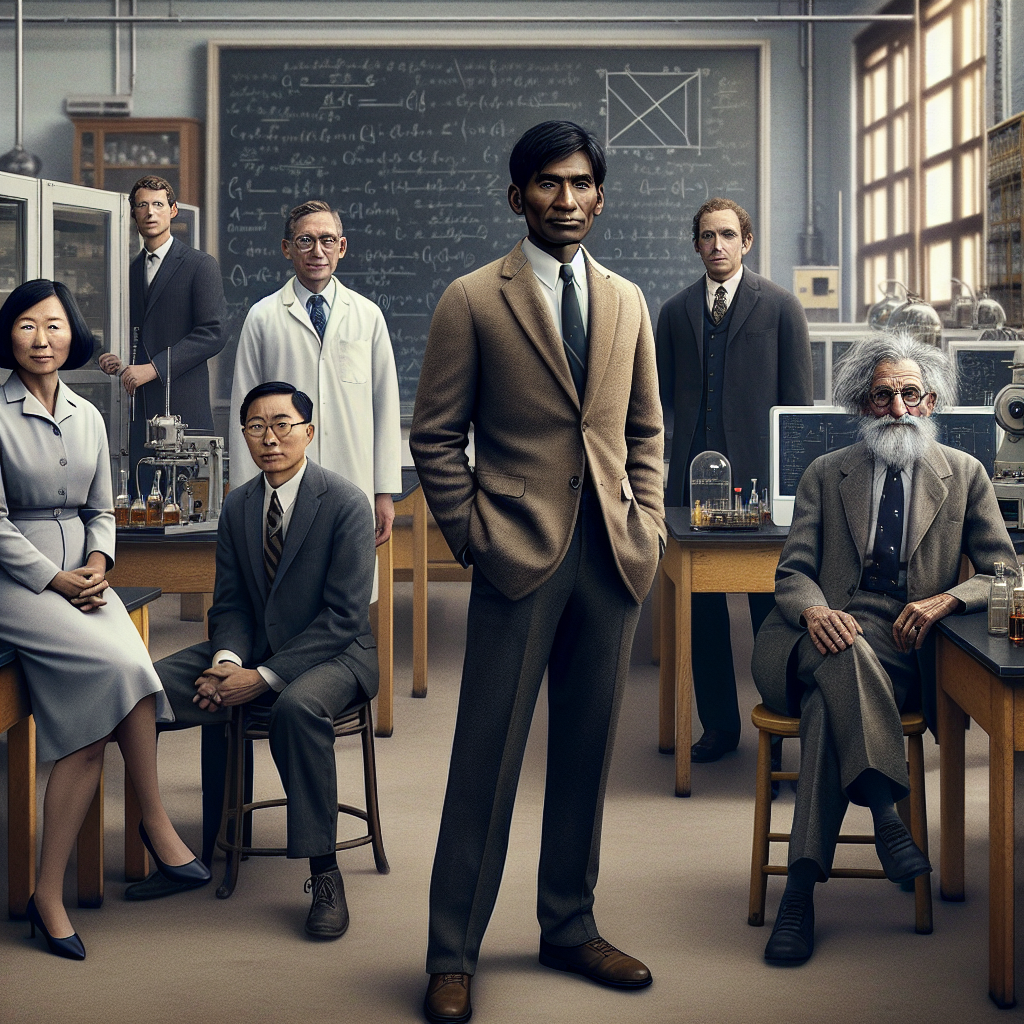}
        \caption*{Iteration 6}
    \end{subfigure}

    \caption{Evolution of the scientific group portrait over six iterations. Each image was generated based on feedback describing which scientist to add and where to place them in the composition.}
    \label{fig:image-generation}
\end{figure}

Despite these limitations, the Evaluator module maintained thematic control by repeatedly anchoring the visual expansion in historically grounded, semantically rich feedback. Figure~\ref{fig:image-generation} presents the full sequence of generated portraits.

\subsubsection{Discussion}

This case study illustrates how EASE orchestrates multimodal generation workflows by integrating text-based reasoning and visual content creation. Compared to algorithmic design, visual iteration introduces new challenges: preserving spatial and stylistic consistency, handling non-determinism in image generation, and mitigating context loss due to token constraints.

Despite these, the framework successfully demonstrated an LLM-mediated expansion of visual content across six structured cycles.

\section{Implementation and Deployment}

The EASE framework is implemented as a modular and extensible system, designed for iterative solution synthesis across multiple domains. The core of the framework is publicly available under an open-source license at:

\begin{center}
    \url{https://github.com/TBU-AILab/EASE}
\end{center}

\subsection{Architecture and Core Functionality}

The backend of EASE is developed in Python, emphasizing modularity, transparency, and easy integration with both LLM APIs and custom evaluation logic. Tasks within EASE are executed in parallel, each running in an isolated thread with its own state, to ensure scalability across multiple concurrent optimization pipelines.

Each Task instance encapsulates a complete iteration loop involving:

\begin{itemize}
    \item A Generator module (typically an LLM, such as OpenAI GPT or similar).
    \item An Evaluator module, which applies either quantitative or qualitative assessment metrics.
    \item Feedback synthesis and prompt injection for the next iteration.
\end{itemize}

Communication between the core and external tools is facilitated through a REST API built using the \texttt{FastAPI} framework. This enables asynchronous interaction with the Task lifecycle, submission of user prompts, retrieval of iteration results, and live monitoring of active jobs.

\subsection{Frontend and Visualization Interface}

To support user-friendly interaction with the system, EASE provides a dedicated web-based frontend, \texttt{FrontEASE}, available at:

\begin{center}
    \url{https://github.com/TBU-AILab/frontEASE}
\end{center}

FrontEASE allows users to:

\begin{itemize}
    \item Create and configure new tasks with customizable module pipelines.
    \item Visualize intermediate outputs across iterations (e.g., code, text, images).
    \item Track task metadata, performance metrics, and Evaluator feedback.
    \item Download results or integrate them into external applications.
\end{itemize}

The frontend (FrontEASE) communicates with the backend (EASE) via REST calls and is implemented using modern web technologies for responsiveness, extensibility, and development efficiency.

FrontEASE is developed on a .NET tech stack, powered by .NET9 framework and C\# language. As the FrontEASE itself is based on a Server-Client architecture, the visualization and client interaction are covered using the Blazor WebAssembly framework (https://dotnet.microsoft.com/en-us/apps/aspnet/web-apps/blazor), allowing for the construction of client logic with the usage of C\# interaction handling alongside standard web technologies (HTML, CSS, JavaScript).

The database technology of choice for preserving the FrontEASE state and backing up FrontEASE-only added functionality (such as user/organization management or resource storage) is the PostgreSQL database (https://www.postgresql.org/) accessed by FrontEASE backend through the ORM solution provided by the Entity Framework (https://learn.microsoft.com/en-us/ef/).

\subsection{Deployment and Reproducibility}

To simplify deployment, EASE is fully containerized using Docker. A default Docker setup is provided in the repository, enabling out-of-the-box execution of the full stack—including the core backend, frontend interface, and any preconfigured LLM connections or simulation environments.

This allows the system to be reproducibly launched in both development and production settings with minimal setup effort.

\subsection{Example Deployment}

An example deployment using Docker is provided to simplify setup and experimentation. The EASE backend and the FrontEASE frontend are both containerized for ease of deployment.

\subsubsection{Step-by-step guide:}

\begin{enumerate}
    \item Clone the repository with submodules:
    \begin{verbatim}
git clone --recurse-submodules https://github.com/TBU-AILab/frontEASE.git
cd frontEASE
    \end{verbatim}

    \item Create a \texttt{.env} file in the root directory with the following contents:
    \begin{verbatim}
# Frontend database default user
POSTGRES_USER="postgres"
POSTGRES_PASSWORD="root"
SEED_DB="true"

# Frontend server and core URLs
API_BASE_URL="http://localhost:4000"
PYTHON_BASE_URL="http://localhost:8086"
    \end{verbatim}

    \item Run Docker Compose to start the containers:
    \begin{verbatim}
docker compose up -d
    \end{verbatim}

    \item Visit the frontend at:
    \begin{verbatim}
http://localhost:5235
    \end{verbatim}

    \item Use the default credentials:
    \begin{itemize}
        \item Username: \texttt{BigJoe}
        \item Password: \texttt{root1234}
    \end{itemize}
\end{enumerate}

This will launch both the EASE backend (available on port 8086) and the FrontEASE interface, allowing users to interact with the framework via a graphical web dashboard or programmatically through the FastAPI-based REST interface.

\subsection*{Extensibility and Custom Modules}

The system supports user-defined modules that can be easily added by implementing the required Python interface. Each module is classified into one of several categories (e.g., LLM Connector, Evaluator, Solution, Analysis), and the framework dynamically integrates available implementations at runtime. This design enables rapid experimentation with new generation strategies, custom fitness functions, or domain-specific evaluation metrics.

A detailed developer guide and documentation for module integration is included in the GitHub repository.

\section{Conclusion}
In this work, we introduced EASE (Effortless Algorithmic Solution Evolution)—a modular, open-source framework that leverages the power of large language models to automate the generation and refinement of algorithmic and creative solutions across diverse domains. EASE enables users to define tasks with structured prompts, evaluate outputs using domain-specific logic or additional LLMs, and iteratively evolve solutions with minimal manual intervention.

Through three case studies—iterative storytelling, 2048 game solver development, and multimodal image composition—we demonstrated EASE's flexibility and adaptability. These examples highlight how EASE orchestrates the integration of generation, testing, analysis, and evaluation into a unified, extensible pipeline, promoting both autonomous creative workflows and performance-driven optimization.

Importantly, EASE abstracts away the technical complexity of prompt engineering and generator management, allowing researchers, developers, and educators to focus on their domain-specific objectives. Whether applied to algorithm synthesis, generative art, or interactive experimentation, EASE provides a reproducible, scalable foundation for human-AI co-design.

By making both the backend and frontend (FrontEASE) freely available, EASE fosters open experimentation, educational engagement, and community-driven extension. As the capabilities of generative models continue to grow, frameworks like EASE will be essential in channeling these advancements into robust, transparent, and accessible solution pipelines.

\backmatter

\bmhead{Acknowledgements}

The research presented in this paper was partially supported by the Internal Grant Agency of Tomas Bata University in Zlin under project number IGA/CebiaTech/2023/004, and by resources of the A.I.Lab at the Faculty of Applied Informatics, Tomas Bata University in Zlin (ailab.fai.utb.cz). Additional support was provided by by program “Excellence initiative—research university” for the AGH University of Krakow as well as the ARTIQ project: UMO-2021/01/2/ST6/00004 and ARTIQ/0004/2021.

\newpage

\begin{appendices}

\appendix
\section{Text Generation: Iterative Story Rewriting with LLM Evaluation}\label{secA1}

This appendix contains the four story versions generated during the EASE-driven rewriting of the ``Little Red Riding Hood'' fairytale. Each version reflects one iteration in the system's feedback-guided refinement cycle.

\subsection*{Iteration 1}
\textbf{Story:}

Little Red Riding Hood set off through the forest with her basket of goodies, humming a tune about nice old grandmothers. Suddenly, the Big Bad Wolf leapt out from behind a tree—except he was wearing a pair of thick-rimmed glasses and a very perplexed expression.

“My dear girl,” said the Wolf, adjusting his spectacles, “I believe you dropped this.” He handed her a bright red cape. “You might want to watch your step. You nearly ended up walking right past Grandma’s cottage”—the Wolf pointed—“which is actually that way.”

Stunned, Little Red Riding Hood thanked him and hurried off. Inside Grandma’s cottage, she found something stranger still: the Wolf sipping tea by the fireplace, reciting poetry to Grandma. As it turned out, Wolf had a sophisticated side, and he’d been helping Grandma write her memoirs. The biggest surprise? Grandma and Wolf had joined forces to open a bakery, and they needed a reliable delivery person.

From that day on, Little Red delivered fresh pastries to all the neighboring villages. And whenever curious travelers asked about the Wolf, she simply replied, “Oh, him? He’s Grandmother’s head pastry chef. Care to try some cookies?”

\vspace{0.5em}
\textbf{Feedback:}
Introduce a new and better twist to the original story of Little red riding hood. Last twist evaluation: 
Rating: 7 Suggestion: The twist of the Wolf being a sophisticated, helpful character who becomes a pastry chef is charming and subversive, but it could be strengthened by adding more tension before the reveal. Perhaps Little Red could initially be terrified when she sees the Wolf with Grandma, assuming the worst (as in the original tale), only to have her expectations completely upended when she realizes they're collaborating on memoirs and a bakery. This creates a stronger contrast between expectation and reality. Additionally, adding a hint about why the Wolf changed from his traditional predatory nature (perhaps he was misunderstood all along, or maybe Grandma helped reform him with her kindness) would give the twist more depth and resonance.

\subsection*{Iteration 2}
\textbf{Story:}

Little Red Riding Hood skipped along the woodland path with fresh bread and honey for her Grandma, imagining the Wolf lurking behind every tree. As she approached the cottage, her heart pounded—terrifying growls came from inside!

She raced through the door, basket held high like a weapon… only to freeze at the most unexpected sight. The Wolf was hunched over Grandma’s old rocking chair, holding something tightly in his claws. He was sobbing.

“Don’t come any closer!” Red shouted, trembling. But the Wolf raised his wet, golden eyes, sniffling, and held up a faded diary, its pages filled with flowery handwriting.

“I didn’t hurt your Grandma,” he whimpered, voice thick with tears. “She’s out in the garden gathering herbs. I was just… reading.” He gently turned the worn pages. “She wrote all about me here—how she found me as a pup, injured and alone. She nursed me back to health. Without her, I—”

At that moment, Grandma bustled in, an armful of lavender in hand. “My word, Wolfy,” she scolded good-naturedly, “I told you to go easy on my diaries. You’ll get tears all over them!”

Little Red blinked in astonishment as Grandma handed the Wolf a handkerchief. “Why… why didn’t you tell me he was your friend?” she gasped.

Grandma smiled. “Because everyone jumps to conclusions about wolves. Better to let you see for yourself.”

Relieved—and a bit embarrassed—Red lowered her basket. All her dread melted away. The Wolf had never been after Grandma’s life; he’d been protecting her secret garden… and cherishing her memories.

From then on, whenever Little Red visited Grandma, she brought an extra loaf of bread for Wolfy. And if villagers whispered warning tales about “the Big Bad Wolf,” Red simply smiled and replied, “Bad? Never met him. But if you’re looking for a friend who’ll cry over Grandma’s tender stories and guard her roses, he’s your fellow.”

\vspace{0.5em}
\textbf{Feedback:}
Introduce a new and better twist to the original story of Little red riding hood. Last twist evaluation:
Rating: 8 Suggestion: The twist that the Wolf is actually Grandma's old friend who she saved as a pup is heartwarming and subverts expectations nicely. To make it even stronger, consider adding a layer where the Wolf has been secretly protecting Red on her journeys through the forest all along (explaining why she's never been harmed despite regularly walking alone through dangerous woods). This would create a satisfying "aha" moment where Red realizes the shadowy figure she's been afraid of in the forest was actually her guardian. Additionally, you could hint that Grandma deliberately sent Red alone through the woods knowing the Wolf would watch over her - a test of both the Wolf's loyalty and Red's courage that ties the generational relationship together more completely.

\subsection*{Iteration 3}
\textbf{Story:}

Little Red Riding Hood walked the forest path alone, her crimson cape bright as a flame in the shady green. Every so often, she sensed a presence behind the trees—soft rustling, gentle eyes glinting in the dark. She hurried on, never quite seeing who (or what) followed her.

At Grandma’s cottage, she felt immense relief… until Grandpa’s old rifle, which hung above the mantel, went off with a startling bang. Startled, Red rushed outside and spotted a wolf sprinting away, something clutched in his mouth. Heart hammering, she chased him into a clearing. Ready for the worst, she confronted him.

Except instead of snarling, the Wolf cautiously approached, dropping a delicate bundle of wildflowers at her feet. “These were Grandma’s favorites,” he said softly. His voice was hoarse and trembling with emotion.

Red blinked in astonishment. “You can talk?”

He nodded. “Your Grandma once saved my life when I was just a pup. I’ve been watching over you ever since you started walking these woods. Didn’t you ever notice that you’ve never been harmed on your journeys here?”

Suddenly, Red remembered the many times she had felt safe when stepping on shadowy paths and half-suspected someone was guiding her. “That was… you?” she asked hesitantly.

“I guarded you from real dangers,” the Wolf whispered. “Even when you were too scared to see, I was there.”

Before Red could respond, Grandma appeared, smiling. “I was testing both of you,” she confessed, hugging them both at once. “You, to see if your bravery could match your kindness… and Wolf, to ensure that your protective spirit is as loyal as ever.”

Red realized then: the forest’s spine-tingling rustles, those golden eyes watching from a distance, had never meant her any harm. They were a silent promise to keep her safe. And so, the cloak she wore wasn’t just a beacon for danger—it was a symbol of trust and unsaid gratitude between a granddaughter, a grandmother, and an unexpected guardian.

\vspace{0.5em}
\textbf{Feedback:}
Introduce a new and better twist to the original story of Little red riding hood. Last twist evaluation:
Rating: 7 Suggestion: The twist showing the wolf as a protective guardian rather than a threat is creative and heartwarming, effectively subverting reader expectations. However, it could be strengthened by addressing a few elements: First, the rifle going off feels disconnected from the plot - why did it fire, and how does this relate to Grandma's "test"? Second, Grandma's revelation that it was all a test feels somewhat contrived. Consider instead having the wolf save Red from a genuine threat (perhaps another predator or human danger), proving his guardianship in action rather than explanation. This would create a more organic revelation of the wolf's true nature and provide a more compelling emotional payoff while maintaining the core subversion of the traditional villain becoming a protector.

\subsection*{Iteration 4}
\textbf{Story:}

With her scarlet hood pulled snug, Little Red Riding Hood made her usual trek to Grandma’s cottage, eyes wary for wolves. Stories of a “Big Bad Wolf” prowled her mind, yet something else skulked in the shadows.

Halfway there, a huntsman suddenly appeared. He tipped his hat, but something in his smile felt cold. “Where’re you headed?” he asked, a hand resting on his gleaming knife.

Red’s heart pounded. He edged closer, knife glinting. Just then, a silent shape lunged from the underbrush—a large wolf. Red shrieked, convinced the beast would pounce on her. Instead, the wolf snarled at the huntsman. A flash of steel, a desperate scuffle. The wolf forced the huntsman to drop his knife and flee.

Shaking, Red expected the wolf to turn on her next. His golden eyes flicked up, meeting her gaze. But he simply laid back his ears and retreated, leaving her breathless but unharmed.

Later, Grandma confirmed the truth: “That wolf has been protecting our cottage for years. The huntsman has hunted beyond the law, and Wolf knows it.”

The chilling rumor turned out to be backward—a dangerous huntsman, a loyal wolf. From then on, Red’s journey remained the same, but her fear of big, bad wolves? Silenced by the echo of a gentle guardian’s howl.

\vspace{0.5em}
\textbf{Feedback:}
Introduce a new and better twist to the original story of Little red riding hood. Last twist evaluation:
Rating: 8 Suggestion: The twist of the wolf being the protector rather than the villain is quite strong and subverts expectations well. To elevate it further, consider adding a personal connection between the wolf and Grandma—perhaps he was once injured and she nursed him back to health, creating a debt of gratitude. Or reveal that the huntsman has been systematically eliminating wolves to monopolize the forest's resources. This would add more emotional resonance and complexity to the twist, making it feel less like a simple role reversal and more like a rich, interconnected story with deeper themes about misjudging others and the true nature of predatory behavior. The revelation could come through Grandma showing Red a scar on the wolf that matches a special bandage only she makes, creating a visual "aha" moment for both Red and the reader. "

\section{Algorithm Generation: 2048 Game Solver via Iterative LLM Refinement}\label{secA2}

This appendix contains the generated Python code of 2048 Game Solvers in 5 consecutive iterations.

\definecolor{mygreen}{rgb}{0,0.6,0}
\definecolor{mygray}{rgb}{0.9,0.9,0.9}
\lstset{
  firstnumber=1,
  numbers=left,
  numberstyle=\footnotesize\color{gray},
  backgroundcolor=\color{mygray},
  language=Python,
  basicstyle=\ttfamily\footnotesize,
  breaklines=true,            
  breakatwhitespace=true,     
  frame=single,
  columns=fullflexible,
  keepspaces=true,
  showstringspaces=false,
  tabsize=4,
  keywordstyle=\color{blue},
  commentstyle=\color{mygreen},
  stringstyle=\color{red}
}

\subsection*{Iteration 1}
\textbf{Solver code:}

\begin{lstlisting}
import numpy as np

def move(grid: np.array, score: int) -> str:

    #Function that determines which direction to move the 2048 grid
    #Output: direction - one of possible moves 'left', 'right', 'up' or 'down'

    # ------------------ [Algorithm body] ------------------

    # Helper function to compress (shift) and merge a single row/array to the left
    def shift_and_combine_left(row):
        # Filter out zeros and keep only non-zero values
        condensed = [val for val in row if val != 0]
        merged = []
        skip = False
        for i in range(len(condensed)):
            # Merge adjacent equal tiles
            if skip:
                skip = False
                continue
            if i < len(condensed) - 1 and condensed[i] == condensed[i+1]:
                merged.append(condensed[i] * 2)
                skip = True
            else:
                merged.append(condensed[i])
        # Pad with zeros to match the row length
        merged.extend([0] * (len(row) - len(merged)))
        return merged

    # Apply a "move left" operation to the entire grid
    def move_left(board):
        new_grid = np.zeros_like(board)
        for i in range(4):
            new_grid[i] = shift_and_combine_left(board[i])
        return new_grid

    # Rotate, move left, and rotate back to achieve the other directions
    def move_right(board):
        # Flip left-right, move left, flip back
        flipped = np.fliplr(board)
        moved = move_left(flipped)
        return np.fliplr(moved)

    def move_up(board):
        # Transpose, move left, transpose back
        transposed = board.T
        moved = move_left(transposed)
        return moved.T

    def move_down(board):
        # Transpose, flip, move left, flip back, transpose back
        transposed = board.T
        flipped = np.fliplr(transposed)
        moved = move_left(flipped)
        unflipped = np.fliplr(moved)
        return unflipped.T

    # Heuristic function to estimate desirability of a board
    def evaluate_board(b):
        # Combine several simple heuristics:
        # 1) High values weighted more if placed near corners
        # 2) Favor empty cells
        # 3) Slightly favor monotonic rows/columns if possible

        # Weights to encourage larger tiles in the top-left corner
        weight_matrix = np.array([
            [16,  12,  8,  6],
            [12,  10,  6,  4],
            [ 8,   6,  4,  2],
            [ 6,   4,  2,  1]
        ], dtype=np.float64)

        # Weighted sum of tiles
        weighted_sum = np.sum(b * weight_matrix)

        # Number of empty cells
        empties = np.count_nonzero(b == 0) * 10

        # Monotonicity check (rows + columns)
        # Simple approximation: penalize adjacent "breaks" in sequence
        monotonic_score = 0
        for row in b:
            for x in range(3):
                if row[x] >= row[x+1]:
                    monotonic_score += 1
        for col in b.T:
            for x in range(3):
                if col[x] >= col[x+1]:
                    monotonic_score += 1

        # Combine heuristics
        return weighted_sum + empties + monotonic_score

    # Try all four moves, pick the one giving the best heuristic result
    directions = ['left', 'right', 'up', 'down']
    best_score = float('-inf')
    best_move = 'left'  # default

    for d in directions:
        if d == 'left':
            new_grid = move_left(grid)
        elif d == 'right':
            new_grid = move_right(grid)
        elif d == 'up':
            new_grid = move_up(grid)
        else:  # 'down'
            new_grid = move_down(grid)

        # If the grid changes, evaluate
        if not np.array_equal(new_grid, grid):
            h = evaluate_board(new_grid)
            if h > best_score:
                best_score = h
                best_move = d

    direction = best_move
    # ------------------ [End of Algorithm body] ------------------

    return direction
\end{lstlisting}

\subsection*{Iteration 2}
\textbf{Solver code:}

\begin{lstlisting}
import numpy as np
import time

def move(grid: np.array, score: int) -> str:
    #Function that determines which direction to move the 2048 grid
    #Output: direction - one of possible moves 'left', 'right', 'up' or 'down'

    # ------------------ [Algorithm body] ------------------
    #
    # This improved solver uses a limited-depth expectimax approach with a simple heuristic.
    # It simulates possible moves ("player nodes") and subsequent tile spawns ("chance nodes")
    # to choose the direction yielding the highest expected outcome.

    # --------------------------------------------------------------
    # Move mechanics: shifting + merging to the left, then rotate/flip
    # to obtain other directions (right, up, down)
    # --------------------------------------------------------------

    def shift_and_combine_left(row):
        merged = []
        nonzero = [x for x in row if x != 0]
        skip = False
        for i in range(len(nonzero)):
            if skip:
                skip = False
                continue
            if i < len(nonzero) - 1 and nonzero[i] == nonzero[i+1]:
                merged.append(nonzero[i]*2)
                skip = True
            else:
                merged.append(nonzero[i])
        merged += [0]*(4-len(merged))
        return merged

    def move_left(bd):
        new_bd = np.zeros_like(bd)
        for i in range(4):
            new_bd[i] = shift_and_combine_left(bd[i])
        return new_bd

    def move_right(bd):
        flipped = np.fliplr(bd)
        merged = move_left(flipped)
        return np.fliplr(merged)

    def move_up(bd):
        transposed = bd.T
        merged = move_left(transposed)
        return merged.T

    def move_down(bd):
        transposed = bd.T
        flipped = np.fliplr(transposed)
        merged = move_left(flipped)
        unflipped = np.fliplr(merged)
        return unflipped.T

    def do_move(bd, direction):
        if direction == 'left':
            return move_left(bd)
        elif direction == 'right':
            return move_right(bd)
        elif direction == 'up':
            return move_up(bd)
        else:  # 'down'
            return move_down(bd)

    # --------------------------------------------------------------
    # Board evaluation
    # --------------------------------------------------------------

    def evaluate_board(bd):
        # Encourage larger tiles in top-left, empties, and mild monotonicity
        weight_matrix = np.array([
            [16,  12,  8,  6],
            [12,  10,  6,  4],
            [ 8,   6,  4,  2],
            [ 6,   4,  2,  1]
        ], dtype=np.float64)
        weighted_sum = np.sum(bd * weight_matrix)
        empties = np.count_nonzero(bd == 0) * 10
        monotonic_score = 0
        for row in bd:
            for x in range(3):
                if row[x] >= row[x+1]:
                    monotonic_score += 1
        for col in bd.T:
            for x in range(3):
                if col[x] >= col[x+1]:
                    monotonic_score += 1
        return weighted_sum + empties + monotonic_score

    # --------------------------------------------------------------
    # Expectimax
    # --------------------------------------------------------------

    # Time limit (avoid going too deep if it takes too long)
    start_time = time.time()
    TIME_LIMIT = 4.9   # a bit below 5 seconds for safety
    MAX_DEPTH = 3

    def expectimax(bd, depth, is_chance):
        # Check time limit or depth => evaluate
        if depth == 0 or (time.time() - start_time) > TIME_LIMIT:
            return evaluate_board(bd)

        if is_chance:
            empty_positions = [(r, c) for r in range(4) for c in range(4) if bd[r][c] == 0]
            if not empty_positions:
                return evaluate_board(bd)
            # For each empty cell, we can spawn 2 (p=0.9) or 4 (p=0.1)
            exp_val = 0.0
            for (r,c) in empty_positions:
                for val, p in [(2, 0.9), (4, 0.1)]:
                    bd[r][c] = val
                    exp_val += p * (expectimax(bd, depth-1, False) / len(empty_positions))
                bd[r][c] = 0
            return exp_val
        else:
            # Player node: choose the max among possible moves
            best_val = float('-inf')
            for direction in ['left','right','up','down']:
                new_bd = do_move(bd, direction)
                if not np.array_equal(new_bd, bd):
                    val = expectimax(new_bd, depth, True)
                    if val > best_val:
                        best_val = val
            # If no valid move changed the board, just evaluate it
            if best_val == float('-inf'):
                return evaluate_board(bd)
            return best_val

    # --------------------------------------------------------------
    # Choose best move by one step of expectimax from the current grid
    # --------------------------------------------------------------

    best_move = 'left'
    best_val = float('-inf')
    for direction in ['left','right','up','down']:
        new_grid = do_move(grid, direction)
        if not np.array_equal(new_grid, grid):
            val = expectimax(new_grid, MAX_DEPTH, True)
            if val > best_val:
                best_val = val
                best_move = direction

    direction = best_move
    # ------------------ [End of Algorithm body] ------------------

    return direction
\end{lstlisting}

\subsection*{Iteration 3}
\textbf{Solver code:}

\begin{lstlisting}
import numpy as np
import time

def move(grid: np.array, score: int) -> str:
    #Function that determines which direction to move the 2048 grid
    #Output: direction - one of possible moves 'left', 'right', 'up' or 'down'

    # ------------------ [Algorithm body] ------------------

    # --------------------------------------------------------------
    # Move mechanics: shift + merge (left), rotate/flip for others
    # --------------------------------------------------------------

    def shift_and_combine_left(row):
        merged = []
        nonzero = [x for x in row if x != 0]
        skip = False
        for i in range(len(nonzero)):
            if skip:
                skip = False
                continue
            if i < len(nonzero) - 1 and nonzero[i] == nonzero[i+1]:
                merged.append(nonzero[i] * 2)
                skip = True
            else:
                merged.append(nonzero[i])
        merged += [0]*(4-len(merged))
        return merged

    def move_left(bd):
        new_bd = np.zeros_like(bd)
        for i in range(4):
            new_bd[i] = shift_and_combine_left(bd[i])
        return new_bd

    def move_right(bd):
        flipped = np.fliplr(bd)
        merged = move_left(flipped)
        return np.fliplr(merged)

    def move_up(bd):
        transposed = bd.T
        merged = move_left(transposed)
        return merged.T

    def move_down(bd):
        transposed = bd.T
        flipped = np.fliplr(transposed)
        merged = move_left(flipped)
        unflipped = np.fliplr(merged)
        return unflipped.T

    def do_move(bd, direction):
        if direction == 'left':
            return move_left(bd)
        elif direction == 'right':
            return move_right(bd)
        elif direction == 'up':
            return move_up(bd)
        else:  # 'down'
            return move_down(bd)

    # --------------------------------------------------------------
    # Heuristic: combine empties, smoothness, monotonicity, weighting
    # --------------------------------------------------------------

    def evaluate_board(bd):
        # Weighted matrix to encourage large tile in corners
        weight_matrix = np.array([
            [16,  12,  8,  4],
            [12,  10,  6,  3],
            [ 8,   6,  4,  2],
            [ 4,   3,  2,  1]
        ], dtype=np.float64)
        weighted_score = np.sum(bd * weight_matrix)
        
        # Count empty cells
        empties = np.count_nonzero(bd == 0)

        # Smoothness: measure of how similar adjacent tiles are
        smoothness = 0
        for r in range(4):
            for c in range(3):
                smoothness -= abs(bd[r][c] - bd[r][c+1])
        for c in range(4):
            for r in range(3):
                smoothness -= abs(bd[r][c] - bd[r+1][c])

        # Monotonicity: reward non-increasing row/column sequences
        monotonic_score = 0
        for row in bd:
            for x in range(3):
                if row[x] >= row[x+1]:
                    monotonic_score += 1
        for col in bd.T:
            for x in range(3):
                if col[x] >= col[x+1]:
                    monotonic_score += 1

        # Combine multiple factors
        return (weighted_score
                + (empties * 20)
                + smoothness/16    # scale smoothness
                + monotonic_score * 2)

    # --------------------------------------------------------------
    # Expectimax with iterative deepening (to stay within time)
    # --------------------------------------------------------------

    TIME_LIMIT = 4.8
    start_time = time.time()

    # Evaluate a node's expected value (chance or player)
    def expectimax(bd, depth, is_chance):
        # If we're out of time or reached depth limit => evaluate
        if (time.time() - start_time) > TIME_LIMIT or depth == 0:
            return evaluate_board(bd)

        if is_chance:
            empty_positions = [(r, c) for r in range(4) for c in range(4) if bd[r][c] == 0]
            if not empty_positions:
                return evaluate_board(bd)
            exp_val = 0
            # For each empty cell, tile spawn is 2 (90\%) or 4 (10\%)
            for (r, c) in empty_positions:
                for tile_val, prob in [(2, 0.9), (4, 0.1)]:
                    bd[r][c] = tile_val
                    exp_val += prob * expectimax(bd, depth - 1, False) / len(empty_positions)
                bd[r][c] = 0
            return exp_val
        else:
            # Player's turn: choose best move
            best_val = float('-inf')
            for direc in ['left', 'right', 'up', 'down']:
                new_bd = do_move(bd, direc)
                if not np.array_equal(new_bd, bd):
                    val = expectimax(new_bd, depth - 1, True)
                    if val > best_val:
                        best_val = val
            if best_val == float('-inf'):
                return evaluate_board(bd)
            return best_val

    def find_best_move(bd, depth):
        best_val = float('-inf')
        best_dir = 'left'
        for direc in ['left','right','up','down']:
            new_bd = do_move(bd, direc)
            if not np.array_equal(new_bd, bd):
                val = expectimax(new_bd, depth, True)
                if val > best_val:
                    best_val = val
                    best_dir = direc
        return best_dir, best_val

    # Iterative deepening: start low depth and go higher until time limit
    best_move_so_far = 'left'
    depth = 1
    while depth <= 7:
        if (time.time() - start_time) > TIME_LIMIT:
            break
        move_candidate, val = find_best_move(grid, depth)
        best_move_so_far = move_candidate
        depth += 1

    direction = best_move_so_far

    # ------------------ [End of Algorithm body] ------------------
    return direction
\end{lstlisting}

\subsection*{Iteration 4}
\textbf{Solver code:}

\begin{lstlisting}
import numpy as np
import time

def move(grid: np.array, score: int) -> str:
    #Function that determines which direction to move the 2048 grid
    #Output: direction - one of possible moves 'left', 'right', 'up' or 'down'

    # ------------------ [Algorithm body] ------------------

    # --------------------------------------------------------------
    # Move mechanics: shift + merge (left), rotate/flip for others
    # --------------------------------------------------------------

    def shift_and_combine_left(row):
        merged = []
        nonzero = [x for x in row if x != 0]
        skip = False
        for i in range(len(nonzero)):
            if skip:
                skip = False
                continue
            if i < len(nonzero) - 1 and nonzero[i] == nonzero[i+1]:
                merged.append(nonzero[i] * 2)
                skip = True
            else:
                merged.append(nonzero[i])
        merged += [0]*(4-len(merged))
        return merged

    def move_left(bd):
        new_bd = np.zeros_like(bd)
        for i in range(4):
            new_bd[i] = shift_and_combine_left(bd[i])
        return new_bd

    def move_right(bd):
        flipped = np.fliplr(bd)
        merged = move_left(flipped)
        return np.fliplr(merged)

    def move_up(bd):
        transposed = bd.T
        merged = move_left(transposed)
        return merged.T

    def move_down(bd):
        transposed = bd.T
        flipped = np.fliplr(transposed)
        merged = move_left(flipped)
        unflipped = np.fliplr(merged)
        return unflipped.T

    def do_move(bd, direction):
        if direction == 'left':
            return move_left(bd)
        elif direction == 'right':
            return move_right(bd)
        elif direction == 'up':
            return move_up(bd)
        else:  # 'down'
            return move_down(bd)

    # --------------------------------------------------------------
    # Advanced heuristic function
    # Incorporates weighted corner, empties, smoothness, merges, monotonicity
    # --------------------------------------------------------------

    def evaluate_board(bd):
        # "Snake" or "corner" weighting helps keep big tiles in corners
        # Use a pattern that encourages monotonic merges along left or top edge
        weight_matrix = np.array([
            [ 65536, 32768, 16384,  8192],
            [  4096,  2048,  1024,   512],
            [   256,   128,    64,    32],
            [    16,     8,     4,     2]
        ], dtype=np.float64)
        snake_weight = np.sum(bd * weight_matrix)

        # Count empty cells
        empties = np.count_nonzero(bd == 0)

        # Smoothness: negative sum of differences between adjacent tiles
        smoothness = 0
        for r in range(4):
            for c in range(3):
                smoothness -= abs(bd[r][c] - bd[r][c+1])
        for c in range(4):
            for r in range(3):
                smoothness -= abs(bd[r][c] - bd[r+1][c])

        # Monotonicity: reward non-increasing or non-decreasing lines
        # We'll do a simpler metric that checks each row/column direction
        monotonic_score = 0
        for row in bd:
            if all(row[i] >= row[i+1] for i in range(3)) or all(row[i] <= row[i+1] for i in range(3)):
                monotonic_score += 1
        for col in bd.T:
            if all(col[i] >= col[i+1] for i in range(3)) or all(col[i] <= col[i+1] for i in range(3)):
                monotonic_score += 1

        # Count merges opportunity
        merges = 0
        for r in range(4):
            for c in range(3):
                if bd[r][c] != 0 and bd[r][c] == bd[r][c+1]:
                    merges += 1
        for c in range(4):
            for r in range(3):
                if bd[r][c] != 0 and bd[r][c] == bd[r+1][c]:
                    merges += 1

        # Combine factors (tune constants to emphasize 2048+ performance)
        return (snake_weight
                + empties * 100
                + smoothness * 0.2
                + monotonic_score * 100
                + merges * 10)

    # --------------------------------------------------------------
    # Expectimax with iterative deepening and move ordering
    # --------------------------------------------------------------

    TIME_LIMIT = 4.8
    start_time = time.time()

    # Use basic move ordering: sort directions by a quick shallow eval
    def quick_evaluate_move(bd, direction):
        new_bd = do_move(bd, direction)
        if np.array_equal(new_bd, bd):
            return -1e9  # invalid move => push to last
        return evaluate_board(new_bd)

    # Expectimax
    def expectimax(bd, depth, is_chance):
        if (time.time() - start_time) > TIME_LIMIT or depth == 0:
            return evaluate_board(bd)

        if is_chance:
            # Chance node: new tile (2 or 4)
            empty_positions = [(r, c) for r in range(4) for c in range(4) if bd[r][c] == 0]
            if not empty_positions:
                return evaluate_board(bd)

            exp_val = 0
            for (row, col) in empty_positions:
                for tile_val, prob in [(2, 0.9), (4, 0.1)]:
                    bd[row][col] = tile_val
                    exp_val += prob * expectimax(bd, depth - 1, False) / len(empty_positions)
                bd[row][col] = 0
            return exp_val
        else:
            # Player node: pick best move
            best_val = float('-inf')
            for direction in ['left','right','up','down']:
                new_bd = do_move(bd, direction)
                if not np.array_equal(new_bd, bd):
                    val = expectimax(new_bd, depth - 1, True)
                    if val > best_val:
                        best_val = val
            if best_val == float('-inf'):
                return evaluate_board(bd)
            return best_val

    def find_best_move(bd, depth):
        # Evaluate possible moves in descending order of quick eval
        directions = ['left', 'right', 'up', 'down']
        directions.sort(key=lambda d: quick_evaluate_move(bd, d), reverse=True)

        best_val = float('-inf')
        best_dir = 'left'
        for direc in directions:
            new_bd = do_move(bd, direc)
            if not np.array_equal(new_bd, bd):
                val = expectimax(new_bd, depth, True)
                if val > best_val:
                    best_val = val
                    best_dir = direc
        return best_dir, best_val

    # Iterative deepening up to a certain depth
    best_move_so_far = 'left'
    depth = 1
    max_depth = 10  # attempt deeper search for better outcomes
    while depth <= max_depth:
        if (time.time() - start_time) > TIME_LIMIT:
            break
        current_best_dir, _ = find_best_move(grid, depth)
        best_move_so_far = current_best_dir
        depth += 1

    direction = best_move_so_far

    # ------------------ [End of Algorithm body] ------------------
    return direction
\end{lstlisting}

\subsection*{Iteration 5}
\textbf{Solver code:}

\begin{lstlisting}
import numpy as np
import time
import random

def move(grid: np.array, score: int) -> str:
    #Function that determines which direction to move the 2048 grid
    #Output: direction - one of possible moves 'left', 'right', 'up' or 'down'

    # ------------------ [Algorithm body] ------------------

    # --------------------------------------------------------------
    # Move mechanics: shift + merge (left), rotate/flip for others
    # --------------------------------------------------------------

    def shift_and_combine_left(row):
        merged = []
        nonzero = [x for x in row if x != 0]
        skip = False
        for i in range(len(nonzero)):
            if skip:
                skip = False
                continue
            if i < len(nonzero) - 1 and nonzero[i] == nonzero[i+1]:
                merged.append(nonzero[i] * 2)
                skip = True
            else:
                merged.append(nonzero[i])
        merged += [0]*(4-len(merged))
        return merged

    def move_left(bd):
        new_bd = np.zeros_like(bd)
        for i in range(4):
            new_bd[i] = shift_and_combine_left(bd[i])
        return new_bd

    def move_right(bd):
        flipped = np.fliplr(bd)
        merged = move_left(flipped)
        return np.fliplr(merged)

    def move_up(bd):
        transposed = bd.T
        merged = move_left(transposed)
        return merged.T

    def move_down(bd):
        transposed = bd.T
        flipped = np.fliplr(transposed)
        merged = move_left(flipped)
        unflipped = np.fliplr(merged)
        return unflipped.T

    def do_move(bd, direction):
        if direction == 'left':
            return move_left(bd)
        elif direction == 'right':
            return move_right(bd)
        elif direction == 'up':
            return move_up(bd)
        else:  # 'down'
            return move_down(bd)

    # --------------------------------------------------------------
    # Heuristic combining "snake weighting", empties, merges, smoothness
    # --------------------------------------------------------------

    def evaluate_board(bd):
        # Weighted matrix to keep large tiles in top-left "snake" pattern
        snake_weights = np.array([
            [ 65536, 32768, 16384,  8192],
            [  4096,  2048,  1024,   512],
            [   256,   128,    64,    32],
            [    16,     8,     4,     2]
        ], dtype=np.float64)
        snake_score = np.sum(bd * snake_weights)

        # Count empty cells
        empties = np.count_nonzero(bd == 0)

        # Smoothness: negative sum of differences between adjacent tiles
        smoothness = 0
        for r in range(4):
            for c in range(3):
                smoothness -= abs(bd[r][c] - bd[r][c+1])
        for c in range(4):
            for r in range(3):
                smoothness -= abs(bd[r][c] - bd[r+1][c])

        # Merges: count potential merges in both horizontal \& vertical
        merges = 0
        for r in range(4):
            for c in range(3):
                if bd[r][c] != 0 and bd[r][c] == bd[r][c+1]:
                    merges += 1
        for c in range(4):
            for r in range(3):
                if bd[r][c] != 0 and bd[r][c] == bd[r+1][c]:
                    merges += 1

        # Combine factors
        return (snake_score 
                + empties * 100
                + merges * 50
                + smoothness * 0.1)

    # --------------------------------------------------------------
    # Expectimax with move ordering + optional Monte Carlo fallback
    # --------------------------------------------------------------

    TIME_LIMIT = 4.8  # a bit under 5s
    start_time = time.time()

    def expectimax(bd, depth, is_chance):
        # If we're out of time or reached depth => evaluate
        if (time.time() - start_time) > TIME_LIMIT or depth == 0:
            return evaluate_board(bd)

        if is_chance:
            # Chance node: tile (2 or 4)
            empties = [(r, c) for r in range(4) for c in range(4) if bd[r][c] == 0]
            if not empties:
                return evaluate_board(bd)
            exp_val = 0
            for (r, c) in empties:
                for val, p in [(2, 0.9), (4, 0.1)]:
                    bd[r][c] = val
                    exp_val += (p * expectimax(bd, depth - 1, False)) / len(empties)
                bd[r][c] = 0
            return exp_val
        else:
            # Player node: pick best move
            best_val = float('-inf')
            directions = ['left', 'right', 'up', 'down']
            # Move ordering: sort by a fast, single-step evaluation
            directions.sort(key=lambda d: evaluate_board(do_move(bd, d)), reverse=True)

            move_found = False
            for direc in directions:
                new_bd = do_move(bd, direc)
                if not np.array_equal(new_bd, bd):
                    move_found = True
                    val = expectimax(new_bd, depth - 1, True)
                    if val > best_val:
                        best_val = val
            if not move_found:
                return evaluate_board(bd)
            return best_val

    # --------------------------------------------------------------
    # Simple random rollout for tie-break or fallback
    # --------------------------------------------------------------

    def random_rollout_score(bd, length=50):
        # Copy board so as not to modify original
        sim_board = np.copy(bd)
        sc = 0
        for _ in range(length):
            # Try a random move
            direction = random.choice(['left','right','up','down'])
            new_bd = do_move(sim_board, direction)
            # If it doesn't change the board => probably stuck, break
            if np.array_equal(new_bd, sim_board):
                break
            sim_board = new_bd
            # 'Spawn' a random tile
            empties = [(r, c) for r in range(4) for c in range(4) if sim_board[r][c] == 0]
            if not empties:
                break
            (r, c) = random.choice(empties)
            sim_board[r][c] = 2 if random.random() < 0.9 else 4
        sc = evaluate_board(sim_board)
        return sc

    # --------------------------------------------------------------
    # Choose best move with a shallow expectimax + random rollout fallback
    # --------------------------------------------------------------

    def find_best_move(bd, depth):
        best_val = float('-inf')
        best_dir = 'left'
        directions = ['left','right','up','down']
        # Move ordering for shallow search
        directions.sort(key=lambda d: evaluate_board(do_move(bd, d)), reverse=True)

        for direc in directions:
            new_bd = do_move(bd, direc)
            if not np.array_equal(new_bd, bd):
                val = expectimax(new_bd, depth, True)
                # Tiebreaker: random rollout
                if abs(val - best_val) < 1e-6:
                    # if nearly equal => do a quick random rollout tiebreak
                    val_tiebreak = random_rollout_score(new_bd)
                    if val_tiebreak > random_rollout_score(do_move(bd, best_dir)):
                        best_dir = direc
                elif val > best_val:
                    best_val = val
                    best_dir = direc
        return best_dir

    # Iterative deepening
    max_depth = 6
    direction_chosen = 'left'
    depth = 1
    while depth <= max_depth:
        if (time.time() - start_time) > TIME_LIMIT:
            break
        direction_chosen = find_best_move(grid, depth)
        depth += 1

    direction = direction_chosen

    # ------------------ [End of Algorithm body] ------------------
    return direction
\end{lstlisting}




\end{appendices}


\bibliography{EAGLE_will_land}

\end{document}